\documentclass[10pt,letterpaper,compsoc,conference]{iiswc26}

\usepackage{cite}
\usepackage{amsmath,amssymb,amsfonts}
\usepackage{algorithmic}
\usepackage{graphicx}
\usepackage[dvipsnames]{xcolor}
\usepackage[final]{microtype}
\usepackage[italic]{mathastext}
\usepackage{libertine}
\usepackage[T1]{fontenc}
\usepackage{textcomp}
\usepackage[varqu,varl]{zi4}
\usepackage[all]{nowidow}
\usepackage[auth-lg,affil-it]{authblk}
\usepackage[keeplastbox]{flushend}
\usepackage{fancyhdr}

\usepackage{formatting/macros}
\usepackage{formatting/packages}
\usepackage{eso-pic}

\fancypagestyle{firstpage}{
  \fancyhf{}
  
}

\begin{document}

\AddToShipoutPictureFG*{%
  \AtPageUpperLeft{%
    \put(\LenToUnit{\paperwidth-0.4in},\LenToUnit{-0.15in}){%
      \makebox[0pt][r]{\raisebox{-\height}{%
        \includegraphics[height=0.55in]{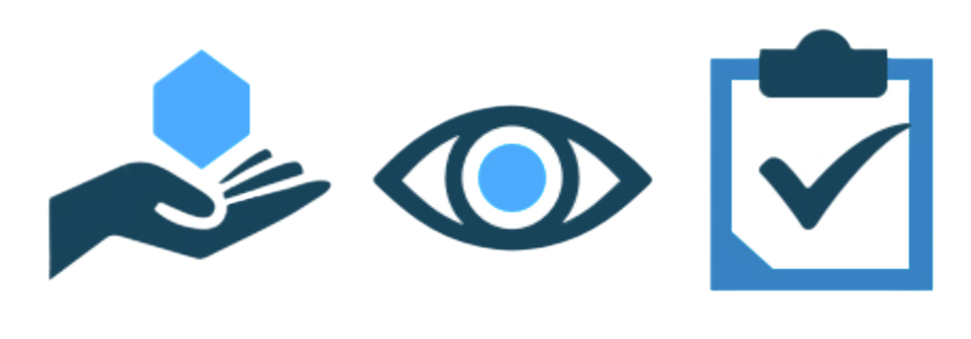}%
      }}}}}


\title{Understanding and Exploiting Diagonal Attention Sparsity in Autoregressive Image Generation}





\author[1]{Daeun Kim}
\author[2]{Junwha Hong}
\author[1]{Changhun Oh}
\author[1]{Yoonsung Kim}
\author[3]{Yoonhyeong Lee}
\author[1]{Jongse Park}
\affil[1]{KAIST}
\affil[2]{Agency for Defense Development}
\affil[3]{Seoul National University}

\maketitle
\thispagestyle{firstpage}
\pagestyle{empty}


\begin{abstract}
Autoregressive image generation has emerged as a paradigm for multimodal AI systems due to its compatibility with transformer-based LLM serving infrastructures. 
However, generating thousands of visual tokens per request makes decoding increasingly bottlenecked by KV cache accesses during attention computation. 
Sparse attention is particularly attractive for this workload because many visual generation applications tolerate moderate quality degradation in exchange for improved performance and efficiency. 
While sparse attention has been extensively explored for text-based LLM inference, it remains unclear whether its sparsity assumptions generalize effectively to autoregressive image generation. 
We present the first systematic characterization of attention sparsity in autoregressive image generation across diverse workloads and representative open-source models. 
Our analysis reveals several distinguishing properties, including a pronounced prefill-decode asymmetry, strong attention concentration on prompt and local tokens, and a unique \textit{diagonal attention sparsity} pattern arising from the spatial locality of visual tokens. 
Motivated by these observations, we propose a diagonal-aware sparse attention mechanism that selectively skips KV entries along the diagonal attention direction within a recent window. 
Implemented on top of a GPU-based serving system using FlexGen, FlashAttention-2, and custom kernels, our approach achieves up to 3.1$\times$ throughput and 1.19$\times$ latency improvements with less than 2\% quality degradation compared to dense inference.

\end{abstract}
\section{Introduction}

Recently, multimodal AI systems are increasingly evolving beyond text generation toward native visual generation capabilities.
Emerging applications such as multimodal agents~\cite{yao2025survey}, interactive image editing~\cite{google2025gemini25flashmodelcard,google2025gemini3proimagemodelcard,gpt4osystemcard}, and simulation-based world models~\cite{agarwal2025cosmos} increasingly rely on generating visual outputs within the model interaction loop.
\textit{Autoregressive image generation} has become a promising paradigm since it formulates visual synthesis as next-token prediction, naturally aligning with transformer architectures and LLM serving infrastructures.
However, autoregressive image generation sequentially generates thousands of visual tokens per request, leaving the decode phase bottlenecked by repeated KV cache accesses in attention computation.

\begin{figure*}[t]
\centering
\includegraphics[width=\linewidth]{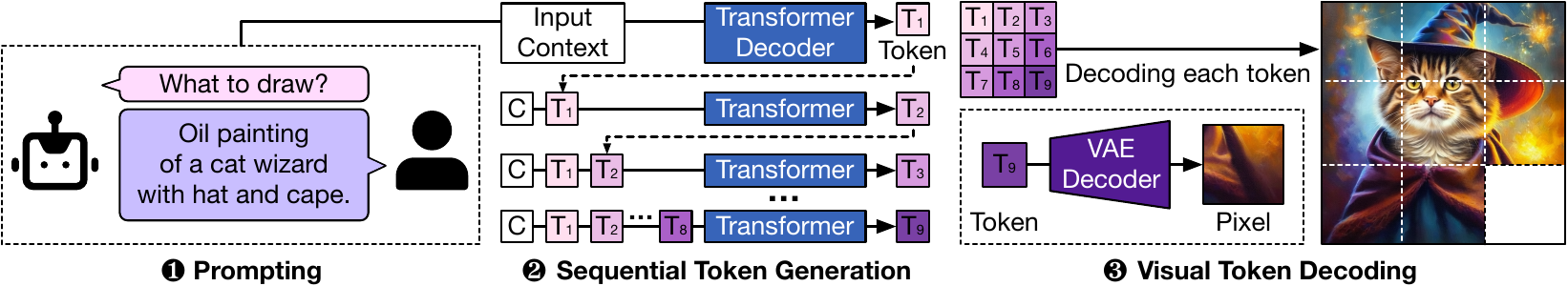}
\caption{Overview of autoregressive image generation pipeline.}
\label{fig:autoregressive_image_generation}
\vspace{-\baselineskip}
\end{figure*}

Generally speaking, for computationally intensive workloads, prior works have extensively explored approximation techniques that trade modest accuracy degradation for substantial gains in performance and efficiency~\cite{han2015deep,zaheer2020big,beltagy2020longformer,ham20203,wang2021spatten,oh2026neo,you2023vitcod,ma2024deepcache,hwang2025d,hwang2022cova}.
Particularly, many visual generation workloads are known to tolerate moderate quality degradation in exchange for improved performance and efficiency{~\cite{park2016axgames, guenter2012foveated,patney2016towards,wang2023foveated}, suggesting that such techniques are attractive for autoregressive image generation.
Practical scenarios such as draft image generation and iterative prompt exploration~\cite{adobe_firefly_fastmode, midjourney_draft} further reinforce this characteristic.
Among these approaches, \textit{sparse attention} has emerged as a promising direction for accelerating autoregressive transformers by selectively attending to only a subset of cached tokens during decoding{~\cite{zhang2023h2o,zhao2024alisa,xiao2023efficient,liu2023scissorhands,oren2024tova,tang2024quest,li2024snapkv,chen2024nacl,cai2024pyramidkv,dong2024less,xiao2025duoattention,hao2025omnikv,feng2026ada}.
However, it remains unclear whether sparse attention techniques originally developed for text-based LLM inference generalize effectively to autoregressive image generation. 
Although both workloads share the same autoregressive transformer decoding framework, visual token generation fundamentally differs from text generation due to the spatial structure and locality inherent in images. 
As a result, the token importance patterns exploited by existing sparse attention techniques may not directly transfer to image generation workloads, requiring a characterization-driven understanding of attention sparsity in autoregressive image generation. 
In particular, two important questions remain unanswered:

\begin{itemize}[leftmargin=12pt]
    \item \textbf{Question 1:} What sparsity structures emerge during autoregressive image generation?
    \item \textbf{Question 2:} How can such structures be exploited for efficient sparse attention?
\end{itemize}

To answer the first question, we systematically characterize the attention sparsity in autoregressive image generation across diverse real-world image-generation workloads and representative open-source models.
Our analysis reveals several distinguishing properties that fundamentally differ from text-based LLM inference.
First, autoregressive image generation exhibits a pronounced prefill-decode asymmetry, where the KV cache is overwhelmingly dominated by decode-phase visual tokens rather than prompt tokens.
Second, attention concentrates on prompt tokens and local visual tokens, making sparse designs that retain them particularly effective.
Most importantly, we observe a structural \textit{diagonal attention sparsity} pattern unique to image generation, where attention importance propagates diagonally across decoding steps due to the spatial locality inherent in visual tokens.
These observations reveal that sparse attention behaviors in autoregressive image generation fundamentally differ from those of text-generation workloads, reversing several design intuitions established in prior sparse attention techniques for LLM inference.
Motivated by these observations, we propose a sparse attention mechanism that explicitly exploits the diagonal attention sparsity observed in autoregressive image generation.
Unlike prior sparse attention techniques for LLM inference that primarily rely on identifying globally important tokens, our approach leverages the observation that attention importance in image generation shifts diagonally across decoding steps.
Based on this insight, we introduce a diagonal-aware sparse attention policy that selectively skips KV entries along the diagonal attention direction within a recent window.
Our design extends sliding-window-style sparse attention with diagonal-aware token selection, enabling more effective sparse decoding for visual token generation.
Importantly, the proposed approach naturally follows from the structural properties identified in our characterization, demonstrating how image-specific sparsity behaviors can guide the design of efficient sparse attention mechanisms for autoregressive image generation.
We implement the proposed sparse attention mechanism on top of a GPU-based autoregressive image generation serving system built upon FlexGen and FlashAttention-2, integrating custom Triton kernels for diagonal-aware sparse attention.
We evaluate the design across diverse open-source autoregressive image generation models and representative text-to-image benchmarks.
Our evaluation shows that the proposed approach consistently achieves a superior quality-performance tradeoff compared to existing sparse attention techniques originally designed for LLM inference.
In particular, diagonal-aware sparse attention preserves generation quality even under highly aggressive sparsity levels, achieving up to 3.1$\times$ throughput improvement and 1.19$\times$ latency improvement with less than 2\% quality degradation compared to dense inference.
These results demonstrate that autoregressive image generation exposes distinct sparsity opportunities fundamentally different from those of text-generation workloads.
More broadly, our work suggests that exploiting workload-specific sparsity structures can substantially improve the efficiency of future GPU-based image generation serving systems without incurring significant quality degradation.

\section{Background and Motivation}

\subsection{Applications of Text-to-Image Generation}

\noindent Recent advances in text-to-image generation have enabled high-fidelity image synthesis from natural language prompts, fueling its adoption across a wide range of services. 
These services impose diverse quality requirements depending on the target application.
While some applications demand high visual fidelity~\cite{hartmann2025genmarketing}, others tolerate moderate quality in exchange for higher throughput or lower latency.
For example, draft generation produces low-resolution previews that let users iterate on prompts before committing to a full-quality generation~\cite{midjourney_draft, adobe_firefly_fastmode}.
Since drafts are intermediate artifacts rather than final deliverables, their fidelity matters far less than the exploration speed.
As another example, safety filtering feeds generated images to a classifier that only needs category-level recognition rather than user-facing quality~\cite{sd_safety_checker, liu2026wukong}.
These quality-tolerant use cases open a design space for aggressive efficiency optimizations, motivating efficient image generation paradigms.

\subsection{Autoregressive Image Generation}
\label{label:auto_image_generation}

\niparagraph{Emergence of autoregressive image generation.}
Earlier image generation models~\cite{betker2023improving, rombach2022high, midjourney} produce high-fidelity outputs. 
However, their architectures are not naturally compatible with existing LLM inference stacks. 
As a result, multimodal AI systems integrate them as standalone generation pipelines, requiring concurrent serving of heterogeneous models.
To overcome this incompatibility, autoregressive (AR) image generation~\cite{sun2024llamagen,chen2025janus,liu2024lumina,wang2024emu3,team2025nextstep,ma2025token} has emerged as a promising direction because it inherits the core modeling principles of LLMs and formulates visual synthesis as next-token prediction, making it naturally compatible with existing LLM inference stacks.
Consequently, frontier AI platforms~\cite{xai2024aurora,google2025gemini3proimagemodelcard,gpt4osystemcard} increasingly adopt AR image generation as an effective integration approach.

\niparagraph{Autoregressive image generation pipeline.}
AR image generation follows sequential visual token decoding, which consists of three stages, as illustrated in Fig.~\ref{fig:autoregressive_image_generation}.
First, the system begins with \circleN{1} \textit{Prompting}, which provides detailed information about the target image.
Second, the system proceeds with \circleN{2} \textit{Sequential token generation}, which iteratively generates each visual token conditioned on the input context and previously generated visual tokens until the full token grid is filled.
Finally, the system performs \circleN{3} \textit{Visual token decoding}, which converts the completed token grid into the output image using a visual decoder, often borrowed from VQ-VAE autoencoders~\cite{van2017neural,razavi2019generating}.

\niparagraph{LLM-aligned decoding behavior.}
AR image generation follows the same decoding process as LLMs.
The decoder transformer iteratively generates tokens through repeated layers of QKV projection, causal self-attention, and Feed-Forward Network.
This LLM-aligned decoding structure allows AR image generation to be analyzed through the lens of LLM inference, opening the door to performance gains from decoding-phase techniques originally proposed for LLMs.
To provide this background, we next review prior work on efficient LLM inference during the decoding phase, with a particular focus on KV cache-related techniques.

\subsection{KV Cache Optimizations for LLM Decoding} \label{sec:bg-kv}

\niparagraph{Decode bottleneck and optimization opportunity.}
During LLM decoding, KV caching reuses the keys and values of previously generated tokens to avoid redundant computation.
However, each decoding step still needs to retrieve the accumulated KV pairs from prior tokens, making KV access a growing overhead as the sequence length increases.
As a result, decoding performance increasingly bottlenecked by the KV cache.
Prior work~\cite{zhang2023h2o, xiao2023efficient, lee2024infinigen, oren2024tova, liu2023scissorhands} has observed that only a subset of cached key-value pairs are critical for generating the token.
This observation has motivated a class of approaches that exploit sparsity in the KV cache by selecting only the relevant pairs at each step, thereby reducing the cost of attending to the KV cache.

\begin{figure}[t]
\centering
\includegraphics[width=\linewidth]{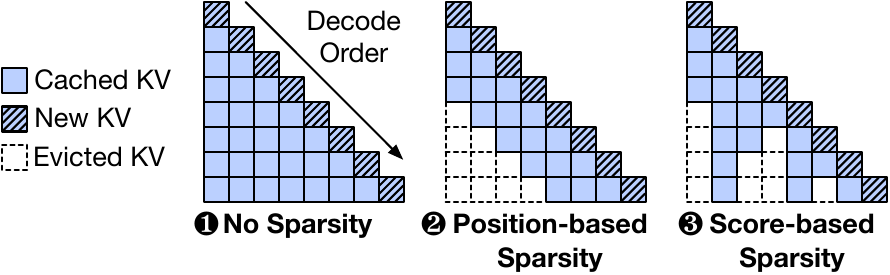}
\caption{Categories of existing KV sparsity techniques.}
\label{fig:kv_cache_pruning}
\vspace{-\baselineskip}
\end{figure}

\niparagraph{KV cache sparsity techniques.}
We categorize KV cache sparsity techniques by their selection criterion for retained KV pairs, as illustrated in Fig.~\ref{fig:kv_cache_pruning}.
\circleNW{1} \textit{No sparsity} serves as the baseline and retains all KV pairs throughout decoding after the prefill stage.
\circleNW{2} \textit{Position-based sparsity} retains KV pairs according to their positions in the cache, yielding retention patterns with positional regularity.
Representative methods include SlidingWindow~\cite{beltagy2020longformer}, which keeps only the most recent KV pairs, and StreamingLLM~\cite{xiao2023efficient}, which augments the local window with a small number of attention-sink tokens (i.e., the first few tokens).
In contrast, \circleNW{3} \textit{Score-based sparsity} selects KV pairs by importance metrics such as attention scores, producing irregular retention patterns scattered across the cache.
These methods implicitly assume that a token's importance persists across subsequent queries, corresponding to vertical patterns in the attention score map.
Methods in this category include H2O~\cite{zhang2023h2o}, TOVA~\cite{oren2024tova}, and ALISA~\cite{zhao2024alisa}, which select KV pairs using accumulated or per-step attention scores.

\niparagraph{Open questions for autoregressive image generation.}
Leveraging existing KV cache sparsity techniques from LLM decoding is a natural starting point for accelerating AR image generation.
However, whether these techniques remain effective under visual-token generation workloads remains an open question.
Unlike text decoding, where token dependencies follow sequential linguistic order, AR image generation produces visual tokens whose attention patterns reflect spatial locality and spatial structure.
These structural differences may fundamentally shift the efficiency-quality tradeoff of KV cache sparsity.
Accordingly, the following sections first characterize autoregressive image generation. 
Building on this analysis, we then revisit existing KV cache sparsity techniques and explore extension directions grounded in visual-token characteristics.


\section{Characterization of Autoregressive Image Generation}
\label{sec:characterization}

\noindent In this section, we characterize autoregressive image generation for sparse KV cache design. 
Section~\ref{subsec:char-methodology} describes our methodology, Section~\ref{subsec:workload} and ~\ref{subsec:bottleneck} analyze sequence lengths and the resulting decoding bottlenecks. 
Then Section~\ref{subsec:attn-sparsity} identifies attention sparsity properties that distinguish AR image generation from text decoding.

\begin{figure}[t]
\centering
\includegraphics[width=\linewidth]{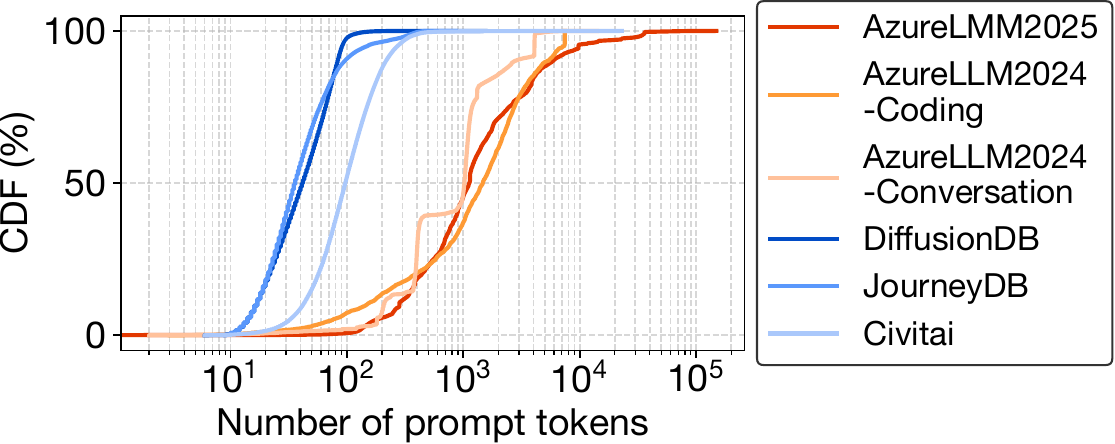}
\caption{Prompt token length distribution in various Text-to-Image and LLM workloads.}
\label{fig:llm-t2i-prompt-len}
\vspace{-\baselineskip}
\end{figure} 

\subsection{Characterization Methodology} \label{subsec:char-methodology}

\noindent We design our methodology to expose workload properties that distinguish autoregressive image generation from text decoding.
To this end, we evaluate representative autoregressive image generation models and LLMs using datasets that reflect their respective serving workloads.
We conduct all experiments on a server-class platform equipped with an NVIDIA RTX A6000 GPU with 48GB of memory and an Intel Xeon Gold CPU.

\niparagraph{Models.}
We analyze three representative open-source autoregressive image generation models: Janus-Pro-1B and 7B~\cite{chen2025janus}, Lumina-mGPT-7B~\cite{liu2024lumina} in resolution 512, and NextStep-1-14B~\cite{team2025nextstep}.
For comparison with text decoding, we include three open-source LLMs of similar scale: Qwen3-8B~\cite{yang2025qwen3}, Llama3.1-8B~\cite{grattafiori2024llama}, and Gemma3-4B~\cite{gemmateam2025gemma3technicalreport}.

\niparagraph{Datasets.}
For image generation, we analyze prompts from three real-world text-to-image traces, including DiffusionDB~\cite{wang2023diffusiondb}, Civitai-prompts~\cite{civitprompts}, and JourneyDB~\cite{sun2023journeydb}, which span Stable Diffusion deployments, Civitai posts, and Midjourney generations, respectively, to capture prompt-length distributions across diverse real-world text-to-image platforms.
For text-generation workloads, we characterize text-generation workloads using three Azure inference traces~\cite{azure_llm_inference}: AzureLMM2025, AzureLLM2024-Coding, and AzureLLM2024-Conversation, which cover multimodal, coding, and conversational serving requests, respectively.

\subsection{Sequence Length Analysis} \label{subsec:workload}

\noindent We first analyze sequence lengths to understand how prefill and decode contribute to KV cache pressure in autoregressive image generation across the workloads.

\begin{table}
\centering
\caption{Output token length of image models.}
\begin{tabular}{cccc} 
\hline
Model                        & Resolution & VAE latent ratio & Output length  \\ 
\hline
Janus-Pro~\cite{chen2025janus}                    & 384x384    & 1/16     & 576            \\ 
\hdashline[1pt/1pt]
Llamagen~\cite{sun2024llamagen}                     & 512x512    & 1/16     & 1024           \\ 
\hdashline[1pt/1pt]
\multirow{2}{*}{Lumina-mGPT~\cite{liu2024lumina}} & 512x512    & 1/16     & 1024           \\
                             & 1024x1024  & 1/16     & 4096           \\ 
\hdashline[1pt/1pt]
\multirow{2}{*}{Emu3~\cite{wang2024emu3}}        & 512x512    & 1/8      & 4096           \\
                             & 720x720    & 1/8      & 8100           \\ 
\hdashline[1pt/1pt]
NextStep-1~\cite{team2025nextstep}             & 512x512    & 1/16     & 1024           \\ 
\hdashline[1pt/1pt]
TokenShuffle~\cite{ma2025token}                 & 2048x2048  & 1/32     & 4096           \\
\hline
\end{tabular}
\label{tab:output-len}
\end{table}

\begin{figure}[t]
\centering
\includegraphics[width=\linewidth]{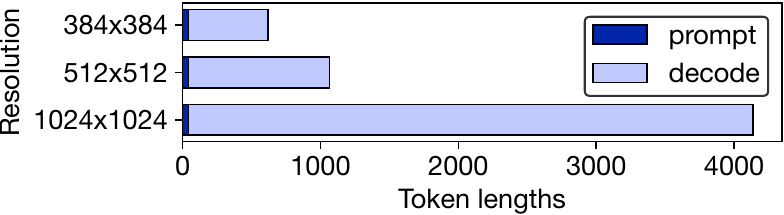}
\caption{Breakdown of prefill and decode token lengths for image model across different target image resolutions.}
\label{fig:t2i-pd-len}
\vspace{-\baselineskip}
\end{figure} 

\niparagraph{Short prompts compared to text generation.}
We analyze prompt token lengths and compare them against representative text generation workloads.
Fig.~\ref{fig:llm-t2i-prompt-len} shows the cumulative distribution function (CDF) of prompt token lengths, with red lines representing LLM workloads and blue lines representing image generation workloads.
Those workloads exhibit much shorter prompt lengths, with most prompts falling below 100 tokens, whereas LLM workloads have average prompt lengths exceeding 1K tokens.
Compared to text generation, where long prompts place substantial pressure on the KV cache, autoregressive image generation workloads incur relatively low prefill-time KV cache pressure due to their short prompt lengths.

\niparagraph{Long decode lengths in image generation.}
In contrast to their short prompts, image generation workloads involve substantially longer decode sequences, whose lengths are fixed by the output resolution and the VAE latent downsampling ratio.
As summarized in Table~\ref{tab:output-len}, most existing models produce roughly 1K to 4K visual tokens under a 1/16 latent ratio, and higher-resolution settings can increase the decode length up to 8100 tokens.
Combined with the prompt-length analysis above, this creates a pronounced prefill-decode asymmetry.
Fig.~\ref{fig:t2i-pd-len} illustrates this asymmetry for models using a typical 1/16 VAE downsampling ratio, where the decode length exceeds the prompt length by up to 40$\times$ at 1024$\times$1024 resolution.
Consequently, visual tokens accumulated during the decode phase dominate the KV cache in autoregressive image generation.

\keyfinding{Sparse KV design should focus on reducing accesses to the long and continuously accumulating decode KV cache, rather than the already short prefill KV cache.}

\subsection{Decoding Phase Bottleneck Analysis} \label{subsec:bottleneck}

\noindent Building on the prefill-decode asymmetry, we next analyze how long visual decode sequences drive memory and compute pressure during decoding.

\begin{figure}[t]
\centering
\includegraphics[width=\linewidth]{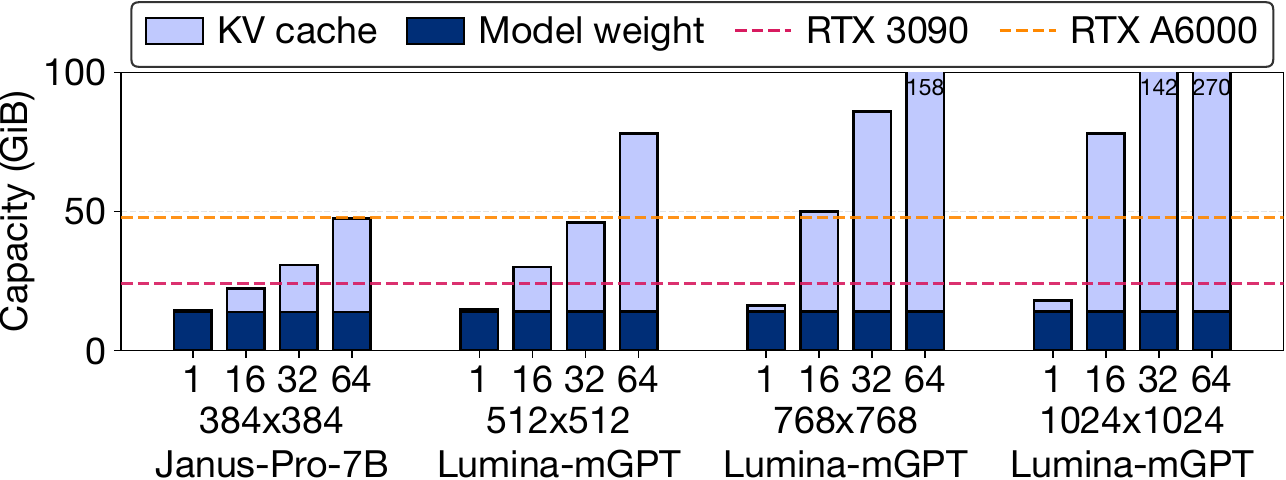}
\caption{Memory footprint in various image generation models. x-axis represents batch size, output image resolution, and model name.}
\label{fig:kv-size}
\end{figure}

\begin{figure}[t]
\centering
\includegraphics[width=\linewidth]{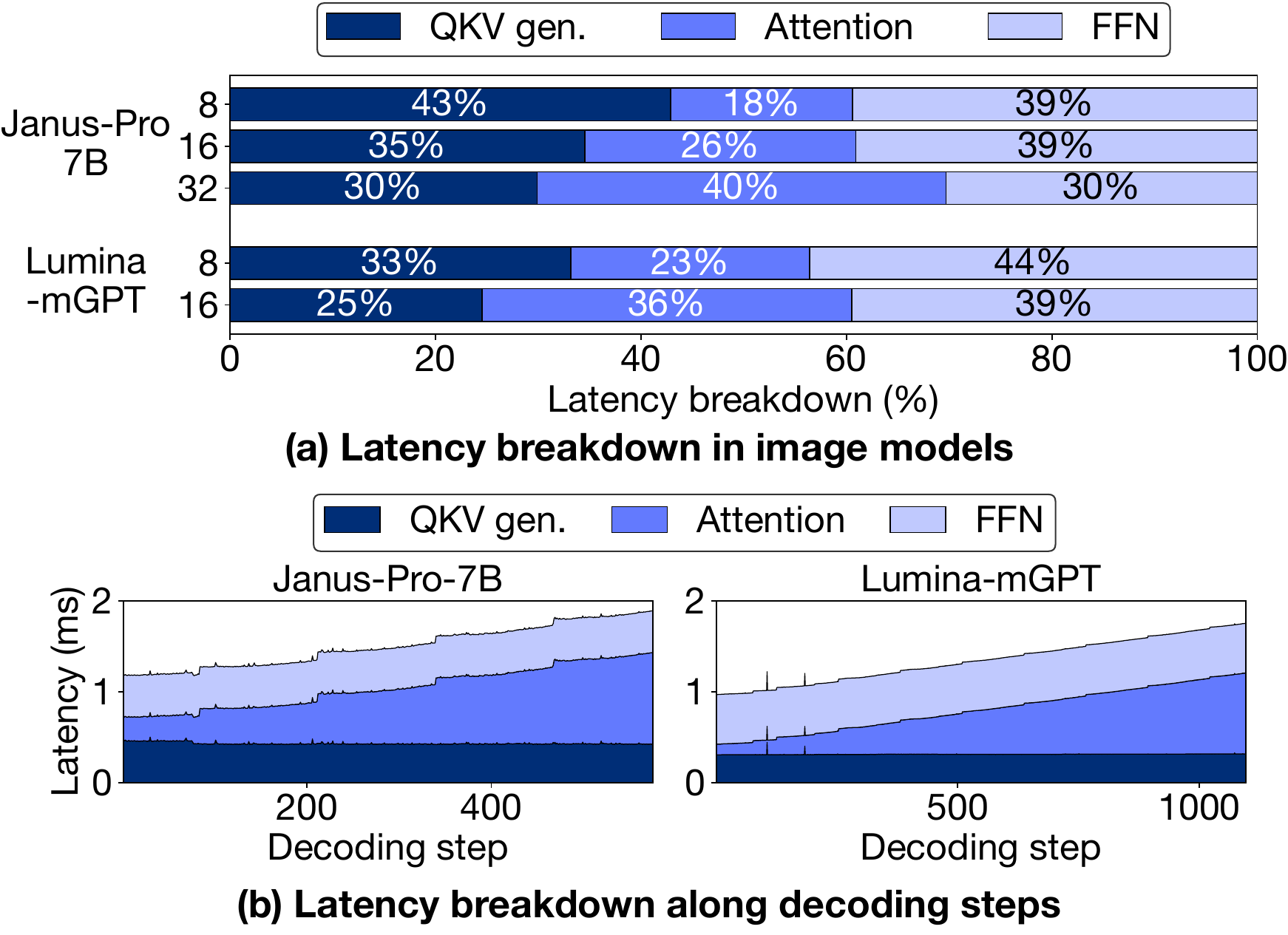}
\caption{Transformer latency breakdown in autoregressive image models. (a) Janus-Pro-7B and Lumina-mGPT latency breakdown across batch sizes, averaged over decoding steps. (b) Per-step latency breakdown for Janus-Pro-7B at batch size 32 (left) and Lumina-mGPT at batch size 16 (right).}
\label{fig:motivation}
\vspace{-\baselineskip}
\end{figure}

\niparagraph{KV cache capacity bottleneck in image generation.}
Image generation services use batched inference to generate multiple images per prompt or improve throughput across concurrent requests.
From this batching perspective, we analyze the per-request memory footprint of representative autoregressive image generation models across resolutions and batch sizes.
As illustrated in Fig.~\ref{fig:kv-size}, we separate model weights from the KV cache and compare the total footprint against RTX 3090 (24GB) and RTX A6000 (48GB) memory budgets.
While lower resolutions allow moderate batching for some models, higher resolutions quickly push the memory footprint beyond practical GPU capacity.
These results show that KV cache capacity becomes a practical bottleneck when autoregressive image generation requires both high batching and high output resolution.

\niparagraph{Attention computation bottleneck during decoding.}
Beyond KV cache capacity, we further examine how different transformer components contribute to end-to-end generation latency.
We break down per-layer latency into three stages using two representative autoregressive image generation models under different batch sizes, as shown in Fig.~\ref{fig:motivation}(a).
The results show that attention accounts for a non-negligible portion of decoding latency, reaching up to 40\%.
To further understand how this overhead evolves during decoding, we analyze the per-step latency breakdown in Fig.~\ref{fig:motivation}(b).
Attention initially accounts for 22.3\% and 12.1\% of per-step latency for Janus-Pro-7B and Lumina-mGPT respectively, but its portion gradually increases to 53.3\% and 50.8\% as decoding progresses.
%
%
Long visual decode sequences not only increase KV cache memory usage but also amplify attention computation, making KV access a central efficiency bottleneck.

\keyfinding{KV cache pressures both memory capacity and attention latency, motivating sparse attention that reduces both axes simultaneously.}

\begin{figure}[t]
\centering
\includegraphics[width=\linewidth]{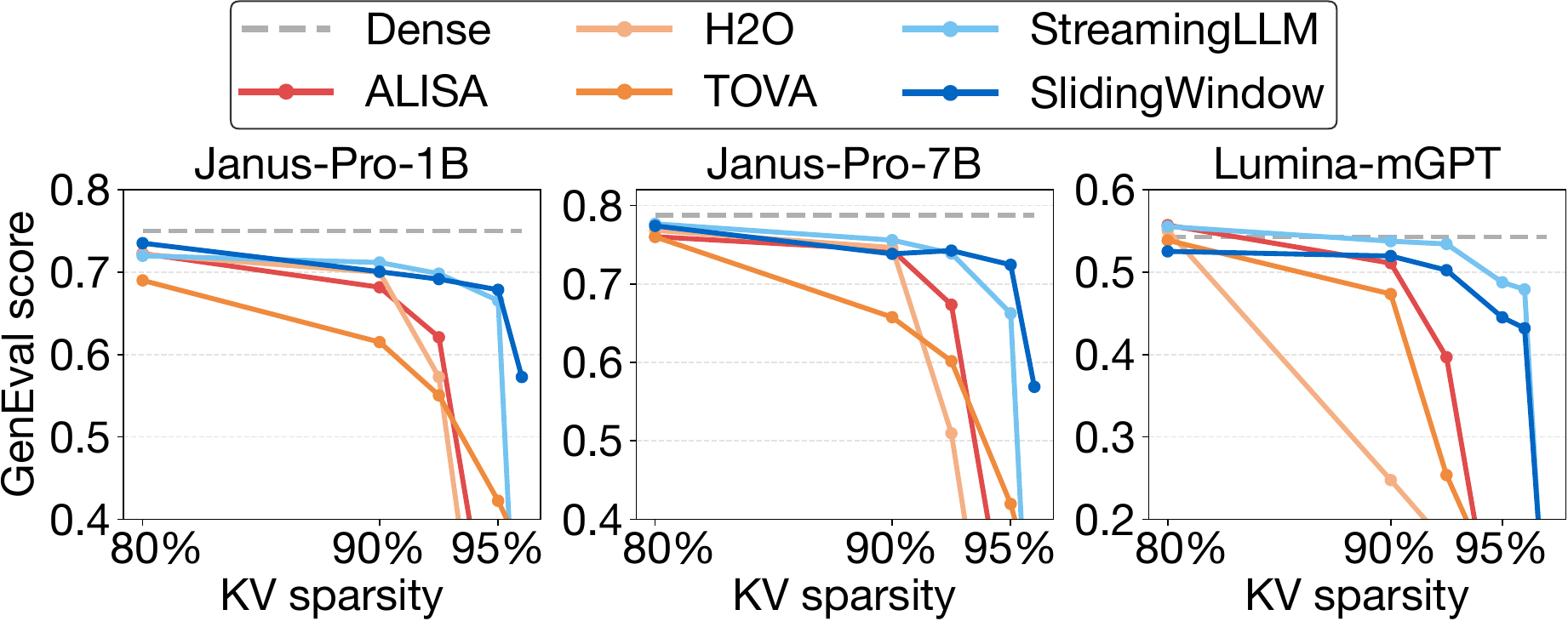}
\caption{GenEval score of LLM baselines applied to image models.}
\label{fig:llm-baseline-acc}
\end{figure}

\begin{figure}[t]
\centering
\includegraphics[width=\linewidth]{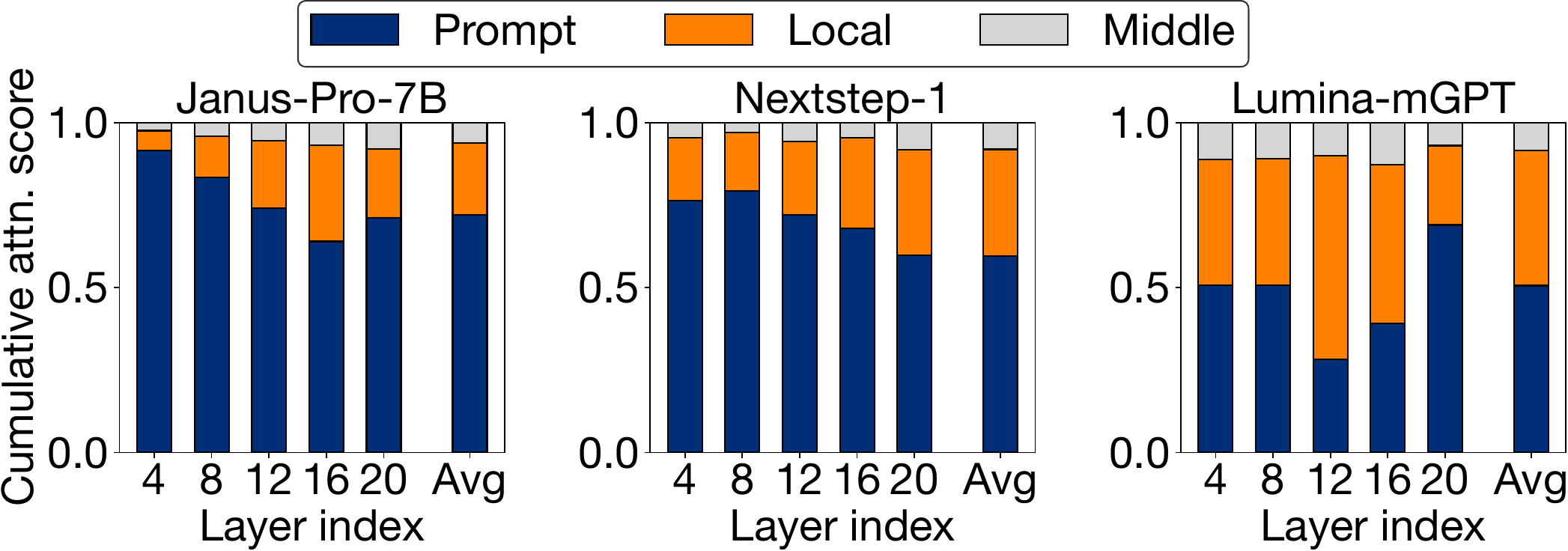}
\caption{Cumulative attention score in prompt, local, and middle token in image models, averaged over all attention heads and decoding steps.}
\label{fig:prompt-window-score}
\vspace{-\baselineskip}
\end{figure}

\subsection{Attention Sparsity Analysis} \label{subsec:attn-sparsity}


\noindent We next characterize attention sparsity as an opportunity to reduce memory footprint and attention cost.
We begin by revisiting existing sparse KV caching methods under AR image generation, and then analyze the attention sparsity underlying their behavior.
 
\niparagraph{Revisiting LLM sparse KV caching works.}
Fig.~\ref{fig:llm-baseline-acc} compares two categories of LLM sparse KV baselines defined in Section~\ref{sec:bg-kv}: blue lines for position-based methods~\cite{zaheer2020big, xiao2023efficient, beltagy2020longformer} and red/orange lines for score-based methods~\cite{zhang2023h2o, oren2024tova, zhao2024alisa}.
Prior LLM studies have shown that score-based selection preserves accuracy better than position-based methods because important tokens are distributed irregularly across the sequence~\cite{zhang2023h2o, tang2024quest}.
In image generation, this ordering reverses: position-based methods consistently outperform score-based methods across all sparsity levels.
We show that this reversal stems from two structural properties of autoregressive image generation: attention concentrates on prompt and recent tokens, and attention sparsity follows a diagonal pattern.
Together, these properties make the irregular middle-token importance patterns exploited by score-based methods in LLMs less effective for image generation workloads.
\keyfinding{Unlike in LLM decoding, position-based sparsity outperforms score-based sparsity in image generation.}

\begin{figure}[t]
\centering
\includegraphics[width=\linewidth]{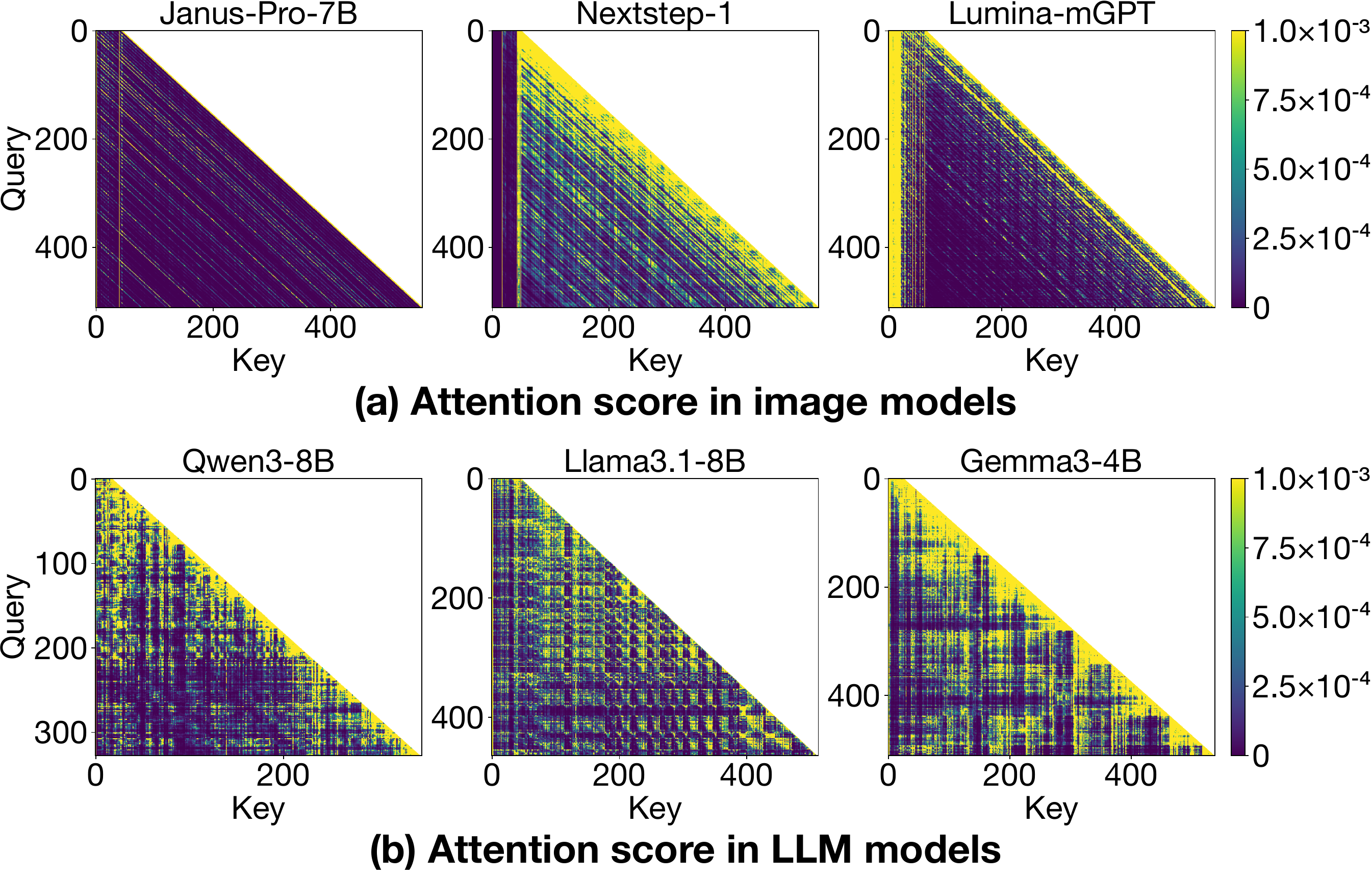}
\caption{Attention score in (a) image models and (b) LLMs. We used attention score map of layer 10 and head 5.}
\label{fig:img-attn-score}
\vspace{-\baselineskip}
\end{figure}

\niparagraph{Skewed attention toward prompt and local tokens.}
Fig.~\ref{fig:prompt-window-score} reports cumulative attention scores over three token groups: prompt tokens, local tokens, defined as the most recent 10\% of decoded tokens, and middle tokens, defined as the remaining decoded tokens.
Although prompt tokens occupy only a small fraction of the full sequence, as shown in Section~\ref{subsec:workload}, they capture more than half of the total attention score.
Local tokens account for most of the attention mass outside the prompt region, while middle tokens contribute only a small fraction.
Across the three models, middle tokens capture only 6.2\%, 8.1\%, and 8.5\% of the total attention score on Janus-Pro, NextStep-1, and Lumina-mGPT.
This skewed pattern reflects the structure of autoregressive image generation. 
Each visual token must remain aligned with the prompt while also preserving spatial consistency with nearby generated tokens.
%

\keyfinding{Attention in image models concentrates on prompt and local tokens, making sparse attention designs that retain them particularly effective.}

\begin{figure}[t]
\centering
\includegraphics[width=\linewidth]{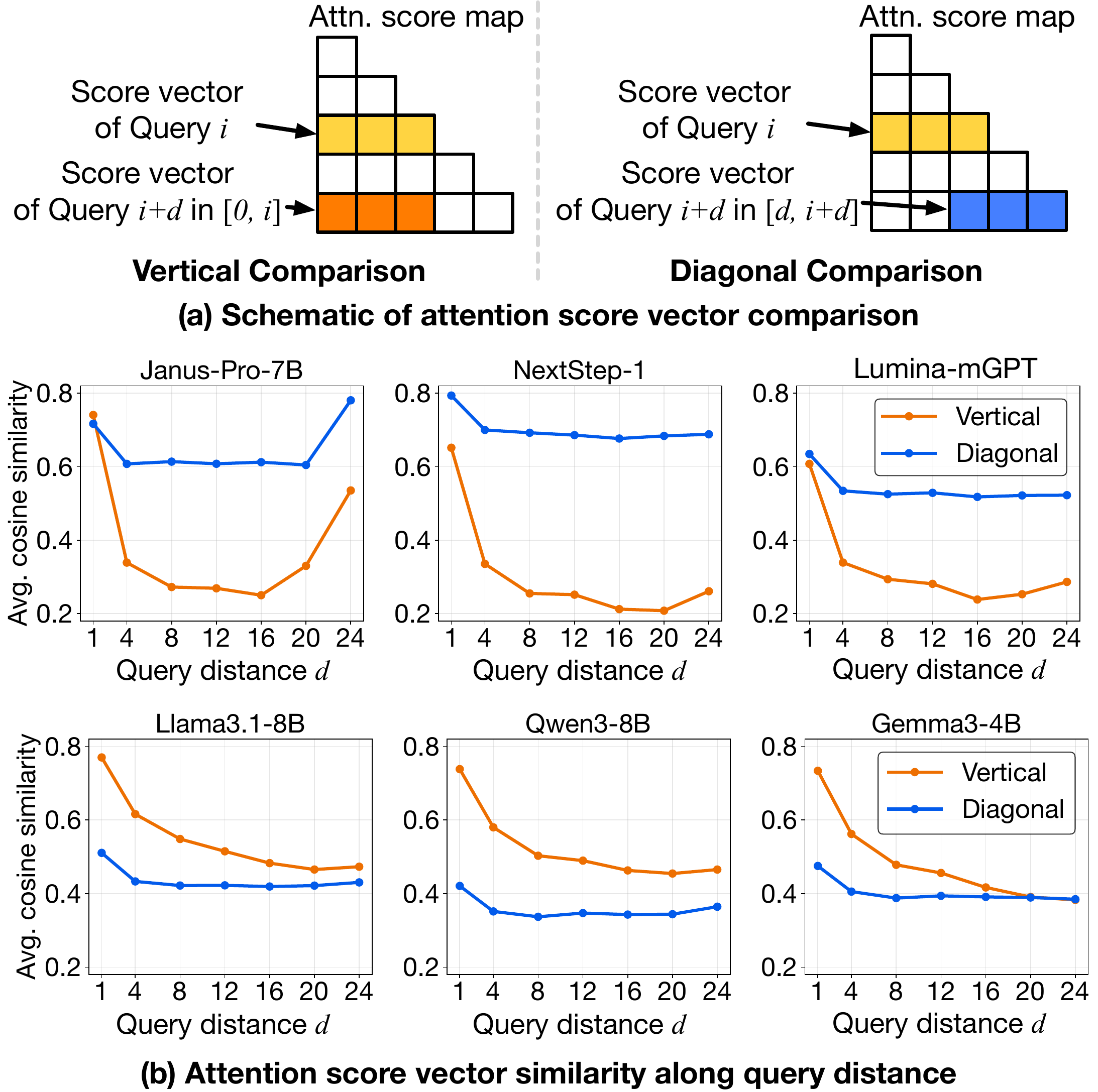}
\caption{(a) Schematic of vertical vs. diagonal attention score vector comparison on a score map. (b) Attention score vector similarity along query distance $d$ for image models(top) and LLMs(bottom), averaged over all layers, heads and queries.}
\label{fig:diag-cosine-similarity}
\vspace{-\baselineskip}
\end{figure}

\niparagraph{Diagonal attention sparsity pattern.}
Fig.~\ref{fig:img-attn-score} visualizes attention scores across three image models and LLMs.
Beyond the prompt region on the left, all three image models exhibit pronounced diagonal stripes spanning the decode region.
In contrast, LLM attention maps show no such structure.
To quantitatively confirm this diagonal structure, we measure the cosine similarity between attention score vectors in two ways, as illustrated in Fig.~\ref{fig:diag-cosine-similarity}(a).
Vertical similarity compares attention score vector of query $i$ and query $i+d$ over their common key range $[0, i]$, capturing how two queries separated by query distance $d$ attend to the same keys.
Diagonal similarity compares attention score vector of query $i$ with query $i+d$ over key range $[d, i+d]$, measuring how similar attention scores are along the diagonal direction of the score map.
As shown in Fig.~\ref{fig:diag-cosine-similarity}(b), diagonal similarity consistently exceeds vertical similarity in image models, contrast to LLMs.
Diagonal alignment is therefore a structural characteristic specific to the image generation.

\niparagraph{Why diagonal patterns emerge.}
In LLMs, attention concentrates on a few globally important tokens, producing vertical patterns in the attention map. 
In image generation, however, no comparably dominant tokens emerge. 
Due to spatial locality, the keys most relevant to each query lie in its neighborhood. 
As the query advances, this neighborhood shifts together with it, preserving the relative position between important query-key pairs. 
And this repeated structure manifests as the diagonal pattern observed in the attention score map shown in Fig.~\ref{fig:diag-reason}.
This confirms that attention in image models is governed by the relative position between query and key, rather than by absolute key positions.
As a result, existing score-based methods fail to capture diagonal sparsity, while conventional position-based methods exploit it only implicitly through local windows.
%
\keyfinding{Spatial locality induces diagonal attention sparsity patterns, motivating sparse attention designs that exploit diagonal structure.}
\vspace{-\baselineskip}

\begin{figure}[t]
\centering
\includegraphics[width=\linewidth]{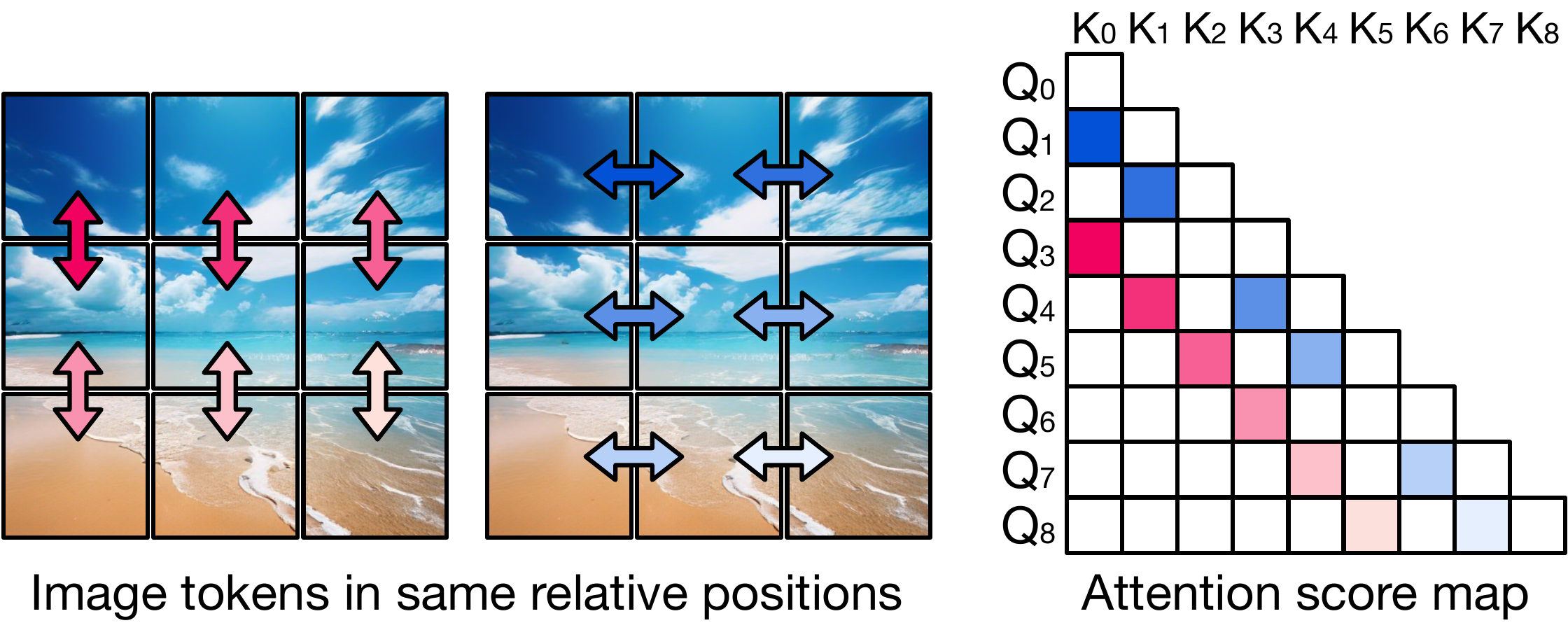}
\caption{Relative position similarity in images projects onto diagonal patterns in the attention score map.}
\label{fig:diag-reason}
\end{figure}

\section{Diagonal-Aware Sparse Attention}
\label{sec:method}
\noindent Our analysis reveals a diagonal attention pattern in autoregressive image generation beyond simple prompt and local-token concentration.
Existing position-based methods partially exploit this through local windows, but degrade sharply at high sparsity.
We therefore propose a sparse attention scheme that explicitly exploits diagonal sparsity to further improve the quality-latency tradeoff.

\niparagraph{Design rationale.}
First, decode KV dominates the cache, so sparsity must target the decode region rather than the prompt. 
Second, attention concentrates on prompt and local tokens while middle tokens contribute negligibly, suggesting that a recent window combined with always-attended prompt KV captures most of the attention weight. 
Third, attention exhibits a diagonal sparsity pattern within the decode region, suggesting that important attention regions can be tracked along diagonal directions rather than simple local windows.

\begin{figure}[t]
\centering
\includegraphics[width=\linewidth]{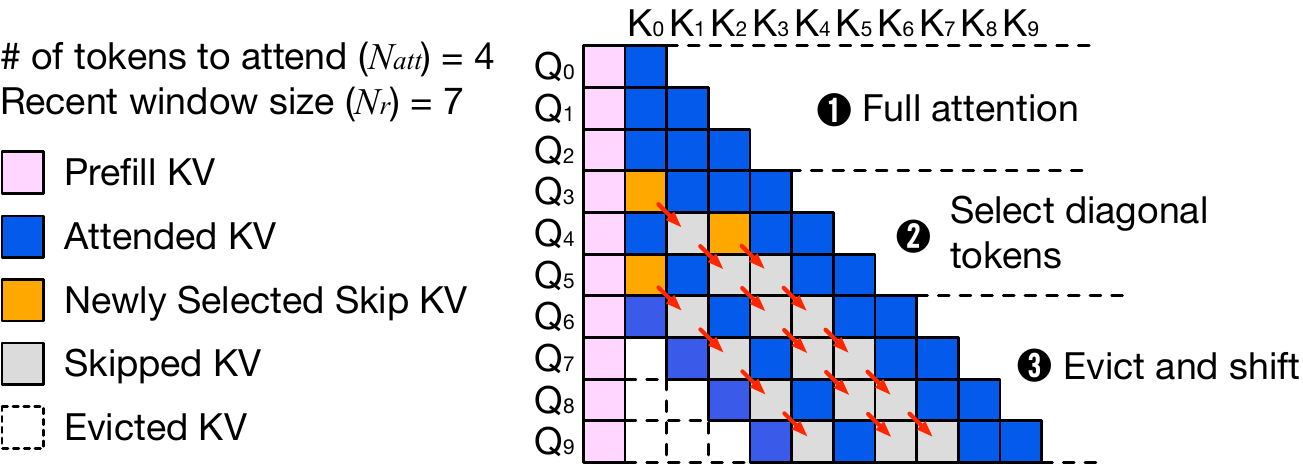}
\caption{Diagonal-aware sparse attention algorithm.}
\label{fig:design}
\vspace{-\baselineskip}
\end{figure}

\begin{algorithm}[t]
\DontPrintSemicolon
\caption{Diagonal-Aware Sparse Attention}
\label{alg:diag-alg}

\KwIn{Recent window size $N_r$}
\KwIn{Number of tokens to attend $N_{att}$}
\KwIn{Decode KV cache $K,V$}
\KwIn{Prompt KV cache $K_p,V_p$}

$skip\_idx \gets [\,]$\;

\For{each decode step $t$}{
    $window \gets [\max(0,t-N_r), \dots, t]$\;

    \uIf{$t < N_{att} - 1$}{
        \tcp{Phase 1: full attention}
        $idx \gets window$\;
        $\mathrm{out}_t \gets
        \mathrm{Attn}(Q_t,\,
        K_p + K[idx],\,
        V_p + V[idx])$\;
    }
    \uElseIf{$t < N_r$}{
        \tcp{Phase 2: grow diagonal skip set}
    
        $skip\_idx \gets [j+1 \;\textbf{for}\; j \in skip\_idx]$\;
    
        $idx \gets window - skip\_idx$\;

        $\mathrm{out}_t \gets
        \mathrm{Attn}(Q_t,\,
        K_p + K[idx],\,
        V_p + V[idx])$\;
        \uIf{$|skip\_idx| < N_r -N_{att}$}{
            \tcp{select newly skipped KV}
            $w \gets \mathrm{softmax}(Q_t \cdot K[idx]^\top)$\;
        
            $j^* \gets \arg\min_j w[j]$\;
        
            $skip\_idx.\mathrm{append}(idx[j^*])$\;
        }
    }
    \Else{
        \tcp{Phase 3: evict oldest KV and shift}
        $K[t-N_r] \gets \varnothing$\;
        $V[t-N_r] \gets \varnothing$\;
        \tcp{Shift diagonal positions by +1}
        $skip\_idx \gets [j+1 \;\textbf{for}\; j \;\textbf{in}\; skip\_idx]$\;
        $idx \gets window - skip\_idx$\;
        $\mathrm{out}_t \gets
        \mathrm{Attn}(Q_t,\,
        K_p + K[idx],\,
        V_p + V[idx])$\;
    }

}
\end{algorithm}

\niparagraph{Algorithm overview.}
Based on the design rationale, our algorithm maintains a wider recent window and computes attention with diagonal token selection. 
Within the recent window, we skip a growing set of positions that trace diagonals through the attention score map, as illustrated in Fig.~\ref{fig:design}.
Once a key receives low attention from $Q_t$, its diagonal neighbor is likely to receive low attention from $Q_{t+1}$, allowing selection decisions to propagate along the diagonal without re-scoring.
We maintain independent diagonal selections for each attention head.

\niparagraph{Detailed explanation.}
Fig.~\ref{fig:design} and Algorithm~\ref{alg:diag-alg} realize these principles with two parameters: recent window size $N_r$ and number of tokens to attend $N_{att}$.
We illustrate with $N_r = 7$ and $N_\text{att} = 4$ as in Fig.~\ref{fig:design}.

\circleN{1} \textbf{Full attention (lines 4-6)} For the first $N_\text{att}-1= 3$ steps, there is no $skip\_idx$ to exclude in attention. 
Each query attends to all available decode KV and prompt KV.

\circleN{2} \textbf{Select diagonal tokens. (lines 7-14)} Once decoding step $t \ge N_\text{att}$, every step appends one new skip KV. 
At $t = 3$, we compute attention scores over $K[0{:}3]$, identify the lowest-scoring key (say $K_0$), and set $skip\_idx$ $= [0]$. 
This newly selected entry is not skipped at the current step.
In the next decoding step ($t=4$), the skip indices shift by $+1$, so $skip\_idx=[1]$, tracking the diagonal successor of $(Q_3,K_0)$.
$Q_4$ then compute attention and score over $\{K_0,K_2,K_3,K_4\}$, select the new lowest (say $K_2$), and append: $skip\_idx= [1, 2]$. 
At $t=5$, the skip indices shift again to $skip\_idx=[2,3]$, continuing to trace the diagonal skip pattern.
The skip set grows by one entry per decoding step until $|skip\_idx|$ becomes $N_r - N_\text{att} (= 3)$.

\circleN{3} \textbf{Evict and shift (lines 15-19)}  Once $t \ge N_r = 7$, the window slides forward and no new keys are selected to be skipped. 
At $t = 7$, the oldest one $K_0$ is evicted from the cache. 
The existing skip indices in $skip\_idx$ shift by $+1$, continuing to trace their diagonals. 
The attended count stays fixed at $N_\text{att} = 4$ throughout this phase.
Throughout all phases, the prompt KV is always attended in full.
%

%
%

\niparagraph{Parameter selection.}
Both parameters $N_r$ and $N_{att}$ are directly determined by the target memory and compute budgets, without per-model tuning.
$N_r$ sets the KV cache footprint and is chosen as a fraction of the expected decoding length $L_{\text{dec}}$; we use $N_r= 0.5 \times L_{\text{dec}}$ in our experiments.
$N_{att}$ sets the per-step attention cost and is chosen to match the desired sparsity ratio.
Together, ($N_r$, $N_{att}$) expose two independent knobs: $N_r$ controls memory, $N_{att}$ controls compute, allowing the user to navigate the memory-compute trade-off directly.

\niparagraph{Cost analysis.}
The KV cache footprint is bounded by $N_r + L_p$, where $L_p$ is the prompt length. 
In contrast, dense attention grows linearly with $L_{\text{dec}} + L_p$.
Per-step attention cost is $O(N_{att} + L_p)$, fixed across decode steps.
The selection step adds an $O(N_r)$ argmin, but it runs only for $N_r - N_{att}$ steps in phase 2 and is amortized over the remaining decoding.
In evict and shift state, the algorithm reduces memory by a factor of $(L_{\text{dec}}+L_p) / (N_r+L_p)$ and per-step compute by $(L_{\text{dec}}+L_p) / (N_{att}+L_p)$ compared to dense attention.
\section{Evaluation}
\subsection{Methodology}
\niparagraph{Implementation.}
We implement our diagonal-aware sparse attention on top of FlexGen~\cite{sheng2023flexgen}, a high-throughput transformer inference engine, integrating FlashAttention-2~\cite{dao2024flashattention} for dense attention and custom Triton~\cite{tillet2019triton} kernels for sparse attention with diagonal selection. 
Our Triton kernel fuses indirect KV gather from the buffer into the attention pass itself, eliminating the intermediate gather buffer. 
It further fuses the argmin over recent-window logits into the attention kernel, removing a separate kernel.

\niparagraph{Models.}
We evaluate on three representative AR image generation models spanning different scales and tokenizer designs. 
Janus-Pro~\cite{chen2025janus} is evaluated at both 1B and 7B parameters with 384$\times$384 resolution (576 visual tokens), and Lumina-mGPT~\cite{liu2024lumina} at 7B parameters with 512$\times$512 resolution (1024 visual tokens).
We set the classifier-free guidance weight to 5.0 for Janus and 4.0 for Lumina-mGPT as default setting in official repository.

\begin{table*}
\centering
\caption{GenEval and DPG-bench score.}
\begin{tabular}{c|c|ccccc|ccccc} 
\hline
\multirow{3}{*}{Model}        & \multirow{3}{*}{Method}   & \multicolumn{5}{c|}{GenEval}                                                       & \multicolumn{5}{c}{DPG-bench}                                                       \\ 
\cline{3-12}
                              &                           & \multicolumn{5}{c|}{Sparsity}                                                      & \multicolumn{5}{c}{Sparsity}                                                        \\
                              &                           & 80\%           & 90\%           & 92.50\%        & 95\%           & 96\%           & 80\%           & 90\%           & 92.50\%        & 95\%           & 96\%            \\ 
\hline
\multirow{8}{*}{Janus-Pro-1B} & Dense                     & 0.750          & -              & -              & -              & -              & 82.22          & -              & -              & -              & -               \\
                              & ALISA                     & 0.723          & 0.682          & 0.621          & 0.192          & 0.118          & 82.17          & 82.06          & 79.65          & 55.07          & 46.47           \\
                              & H2O                       & 0.722          & 0.700          & 0.573          & 0.048          & 0.033          & 82.10          & 81.74          & 72.17          & 39.05          & 36.72           \\
                              & TOVA                      & 0.690          & 0.615          & 0.550          & 0.422          & 0.361          & 81.79          & 79.78          & 76.08          & 72.04          & 68.14           \\
                              & SlidingWindow             & \textbf{0.735} & 0.701          & 0.692          & 0.679          & 0.573          & \textbf{82.86} & \textbf{82.67} & 81.72          & \textbf{81.86} & 77.70           \\
                              & StreamingLLM              & 0.720          & 0.712          & 0.698          & 0.665          & 0.083          & 82.79          & 82.58          & 82.11          & 81.29          & 44.70           \\
                              & \textbf{Diagonal}         & 0.719          & \textbf{0.718} & 0.692          & \textbf{0.688} & 0.668          & 82.04          & 81.51          & \textbf{82.21} & 81.46          & \textbf{81.01}  \\
                              & \textbf{Diagonal + sink4} & 0.716          & 0.710          & \textbf{0.704} & 0.682          & \textbf{0.671} & 81.89          & 81.39          & 81.63          & 81.02          & 80.75           \\ 
\hline
\multirow{8}{*}{Janus-Pro-7B} & Dense                     & 0.788          & -              & -              & -              & -              & 84.16          & -              & -              & -              & -               \\
                              & ALISA                     & 0.760          & 0.743          & 0.674          & 0.230          & 0.125          & 83.46          & 82.83          & 81.07          & 60.91          & 48.83           \\
                              & H2O                       & 0.768          & 0.746          & 0.510          & 0.024          & 0.019          & 83.09          & 83.09          & 73.12          & 37.91          & 34.58           \\
                              & TOVA                      & 0.760          & 0.658          & 0.602          & 0.420          & 0.307          & 82.80          & 80.13          & 78.15          & 73.30          & 67.25           \\
                              & SlidingWindow             & 0.774          & 0.738          & 0.743          & 0.724          & 0.569          & 83.77          & 83.33          & 84.15          & \textbf{83.88} & 78.25           \\
                              & StreamingLLM              & 0.777          & 0.756          & 0.739          & 0.662          & 0.071          & 83.42          & 83.18          & 83.27          & 81.09          & 44.33           \\
                              & \textbf{Diagonal}         & \textbf{0.792} & 0.758          & 0.768          & \textbf{0.749} & 0.738          & 83.55          & \textbf{83.64} & \textbf{83.62} & 83.18          & \textbf{82.67}  \\
                              & \textbf{Diagonal + sink4} & 0.785          & \textbf{0.784} & \textbf{0.774} & 0.743          & \textbf{0.749} & \textbf{83.81} & 83.35          & 82.58          & 82.28          & 82.00           \\ 
\hline
\multirow{8}{*}{Lumina-mGPT}  & Dense                     & 0.543          & -              & -              & -              & -              & 75.67          & -              & -              & -              & -               \\
                              & ALISA                     & \textbf{0.557} & 0.511          & 0.397          & 0.051          & 0.035          & \textbf{76.60} & 74.25          & 67.41         & 46.24          & 42.12           \\
                              & H2O                       & 0.551          & 0.248          & 0.170          & 0.016          & 0.037          & 75.15          & 60.34          & 57.11          & 35.95          & 40.14           \\
                              & TOVA                      & 0.539          & 0.474          & 0.254          & 0.103          & 0.058          & 75.76          & 69.53          & 59.36          & 51.58          & 48.68           \\
                              & SlidingWindow             & 0.525          & 0.520          & 0.502          & 0.445          & 0.432          & 74.88          & 74.13          & 73.17          & 69.35          & 68.46           \\
                              & StreamingLLM              & 0.555          & 0.538          & 0.534          & 0.488          & 0.479          & 75.92          & \textbf{75.64} & 74.57          & 71.43          & 71.37           \\
                              & \textbf{Diagonal}         & 0.542          & 0.540          & 0.547          & 0.520          & 0.492          & 75.93          & 75.15          & 75.13          & \textbf{74.48} & \textbf{72.89}  \\
                              & \textbf{Diagonal + sink4} & 0.555          & \textbf{0.541} & \textbf{0.548} & \textbf{0.524} & \textbf{0.504} & 76.23          & 75.20          & \textbf{75.23} & 73.68          & 72.24           \\
\hline
\end{tabular}
\label{tab:accuracy}
\vspace{-\baselineskip}
\end{table*}

\niparagraph{Sparse KV baselines.}
We compare against dense decoding and five sparse KV baselines.
\textit{Position-based} methods retain KV pairs by positional rules: StreamingLLM~\cite{xiao2023efficient} keeps 4 attention sink tokens together with a recent window, while SlidingWindow~\cite{beltagy2020longformer} keeps only the most recent window.
For StreamingLLM, we retain the first 4 decode tokens as attention sinks, since prompt KV is always kept in full in our setting.
\textit{Score-based} methods retain KV pairs by importance metrics: H2O~\cite{zhang2023h2o} selects heavy hitters by accumulated attention scores, TOVA~\cite{oren2024tova} evicts the minimum-scoring keys at each step, and ALISA~\cite{zhao2024alisa} combines sparsity-aware selection with a recent accumulated scores.
Across all methods, we sweep KV sparsity ratios of 80\% to 96\% relative to the dense decode KV, where the ratio denotes the fraction of decode KV not used for the attention computation.

\begin{figure}[t]
\centering
\includegraphics[width=0.8\linewidth]{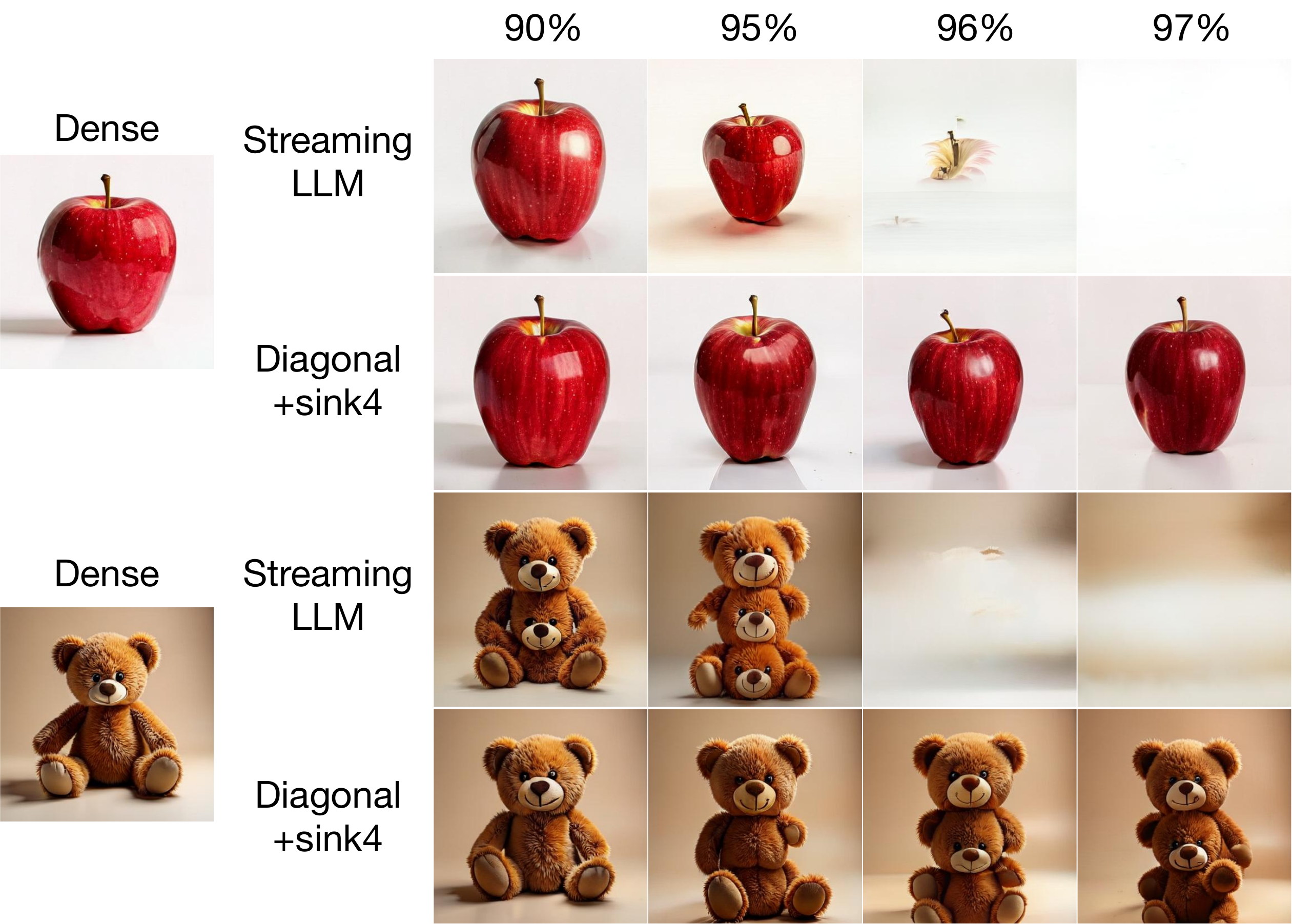}
\caption{Qualitative result in Janus-Pro-7B at high sparsity.}
\label{fig:janus7b-qualitative}
\vspace{-\baselineskip}
\end{figure}

\niparagraph{Our methods.}
We refer to our two variants as \textit{Diagonal} and \textit{Diagonal+Sink4} throughout the evaluation.
Diagonal+Sink4 retains the 4 initial decode tokens as attention sinks, following StreamingLLM, and applies diagonal selection for the remaining slots.
Sparsity ratios refer to the fraction of skipped decode KVs relative to the maximum decode length, i.e., $1 - N_{att} / L_{dec}$.

\niparagraph{Benchmarks.}
We evaluate generation quality on two standard text-to-image benchmarks. 
GenEval~\cite{ghosh2023geneval} measures object-focused prompt alignment across 553 prompts with 4 images generated per prompt, scoring attributes such as object presence, count, color, and spatial relations. 
DPG-Bench~\cite{hu2024ella} complements GenEval with dense, compositional prompts that stress fine-grained semantic alignment. 

\niparagraph{System setup.}
All experiments run on a single NVIDIA RTX A6000 (48GB) with an Intel Xeon Gold CPU. 
We report end-to-end per-image latency and throughput (images per second). 
Latency and throughput are measured with a fixed prompt length of 50 tokens, the median prompt length in DiffusionDB~\cite{wang2023diffusiondb}.
All latency and throughput results are averaged over 3 inferences after a warm-up.

\begin{figure}[t]
\centering
\includegraphics[width=0.8\linewidth]{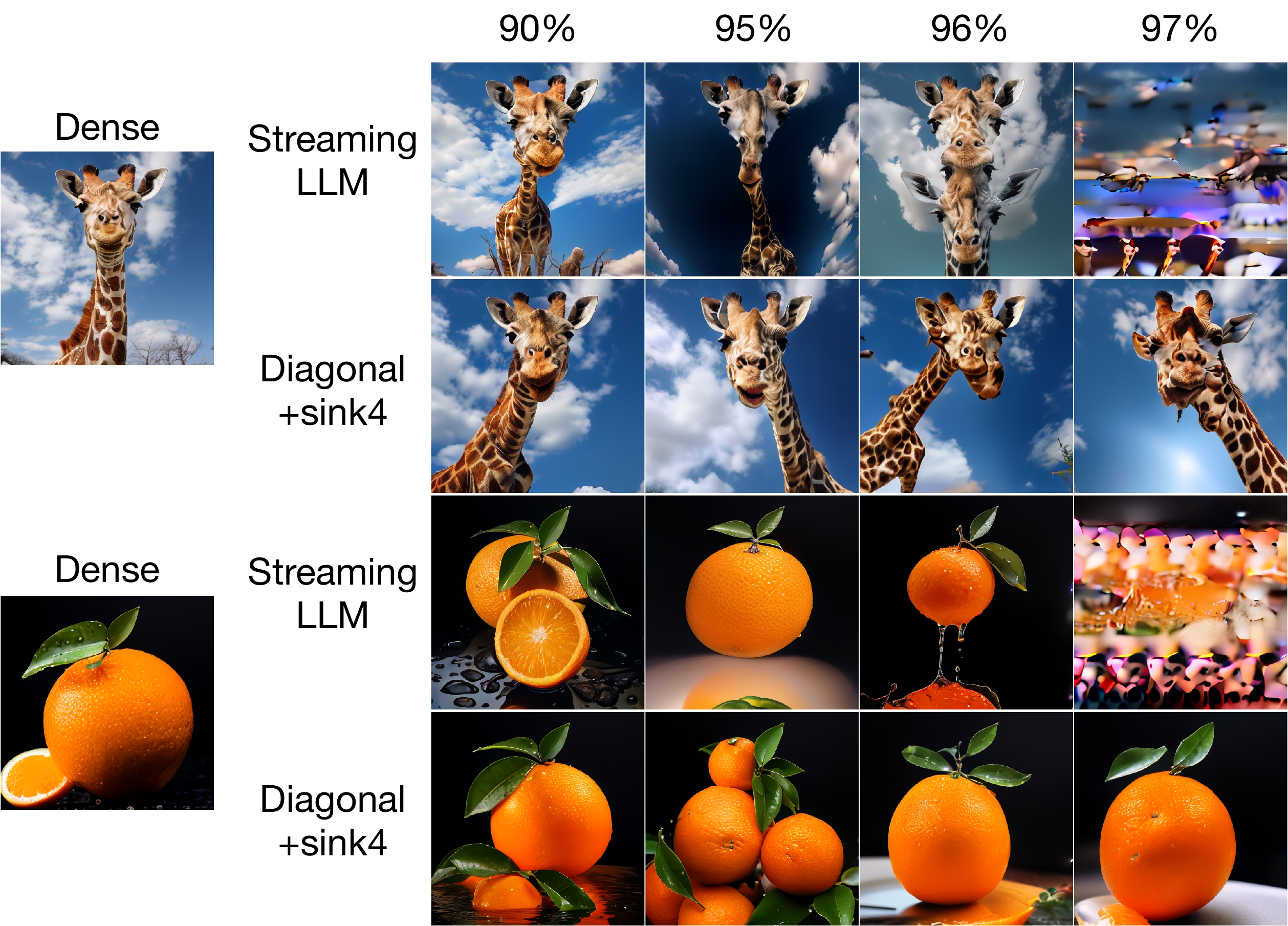}
\caption{Qualitative result in Lumina-mGPT at high sparsity.}
\label{fig:lumina-qualitative}
\vspace{-\baselineskip}
\end{figure}

\subsection{Generation Quality}

\niparagraph{Quantitative results.}
Table~\ref{tab:accuracy} reports GenEval and DPG-bench scores across three models.
First, score-based baselines (ALISA, H2O, TOVA) collapse beyond 90\% sparsity, retaining less than 30\% of dense accuracy at 95-96\%. 
This confirms Takeaway \#3: methods tuned for irregular LLM patterns fail when image attention is dominated by prompt and local structure.
Second, position-based baselines hold up through 90-92.5\% but degrade at 95-96\%. 
StreamingLLM shows a sharp cliff at 96\% on both Janus-Pro models, losing over 90\% of dense accuracy.
Third, Diagonal selection holds the accuracy frontier at extreme sparsity. 
Across all three models, Diagonal stays within 1–6\% of dense accuracy even at 96\% sparsity, where the best position-based baseline drops by 10-30\% and score-based baselines drop by over 80\%. 
Position-based baselines reaches high sparsity by shrinking the recent window itself, restricting attention to a narrow local region. 
Our method instead observes a wider recent window and skips diagonally-aligned attention within it. 
Attention covers the same number of keys drawn from a broader context, which explains the consistent quality gain.

\niparagraph{Qualitative results.}
Fig.~\ref{fig:janus7b-qualitative} and Fig.~\ref{fig:lumina-qualitative} compare StreamingLLM and Diagonal selection against dense outputs.
At 90\% sparsity, both methods preserve image fidelity.
At higher sparsity, StreamingLLM degrades rapidly with repetitive objects and distorted structures.
Diagonal selection retains coherent shapes and prompt-aligned content up to 96-97\%, which is consistent with quantitative results. 

\subsection{Performance}

\begin{figure}[t]
\centering
\includegraphics[width=\linewidth]{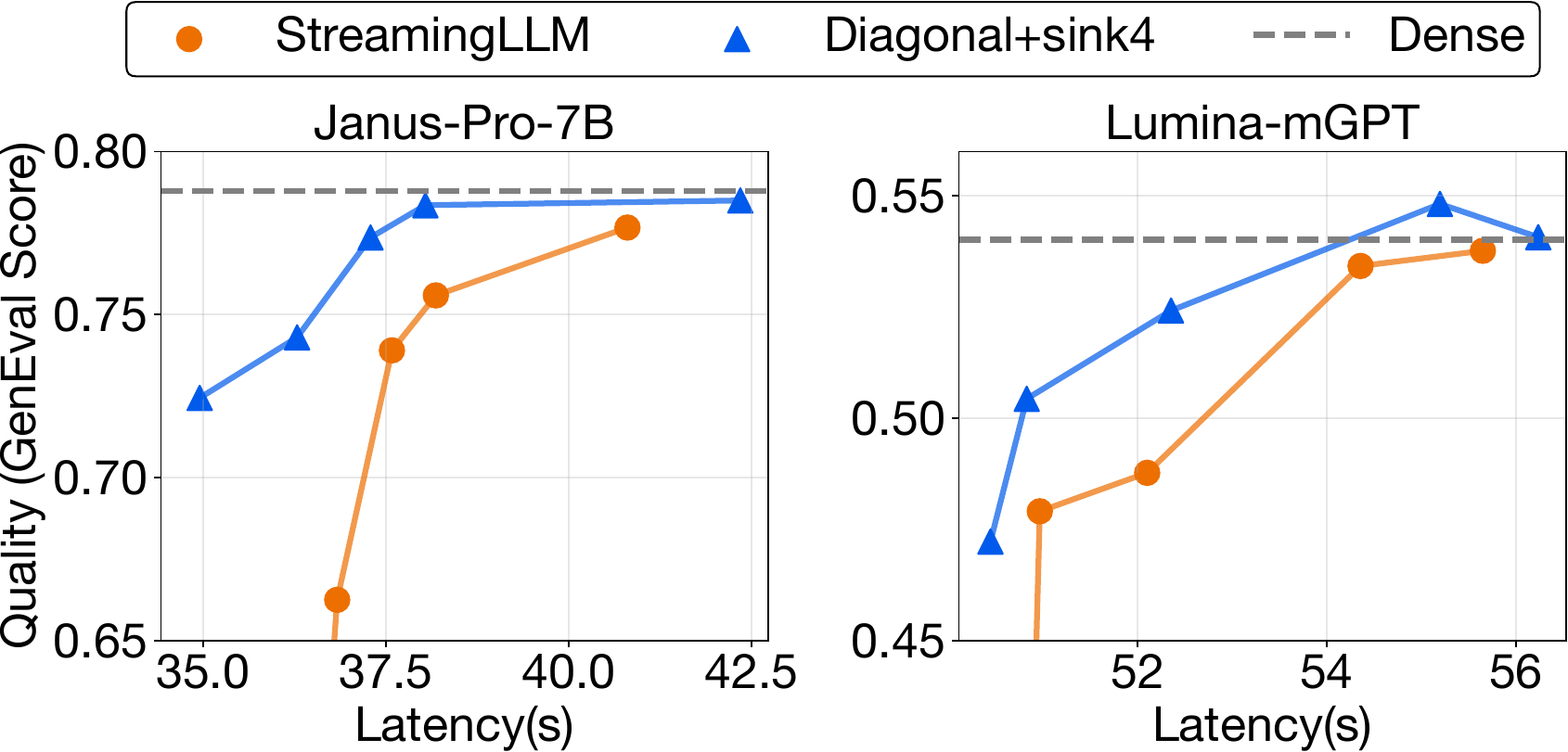}
\caption{Quality-latency tradeoff. Janus-Pro-7B used batch size of 96, and Lumina-mGPT used batch size of 54.}
\label{fig:latency-tradeoff}
\end{figure}

\begin{figure}[t]
\centering
\includegraphics[width=\linewidth]{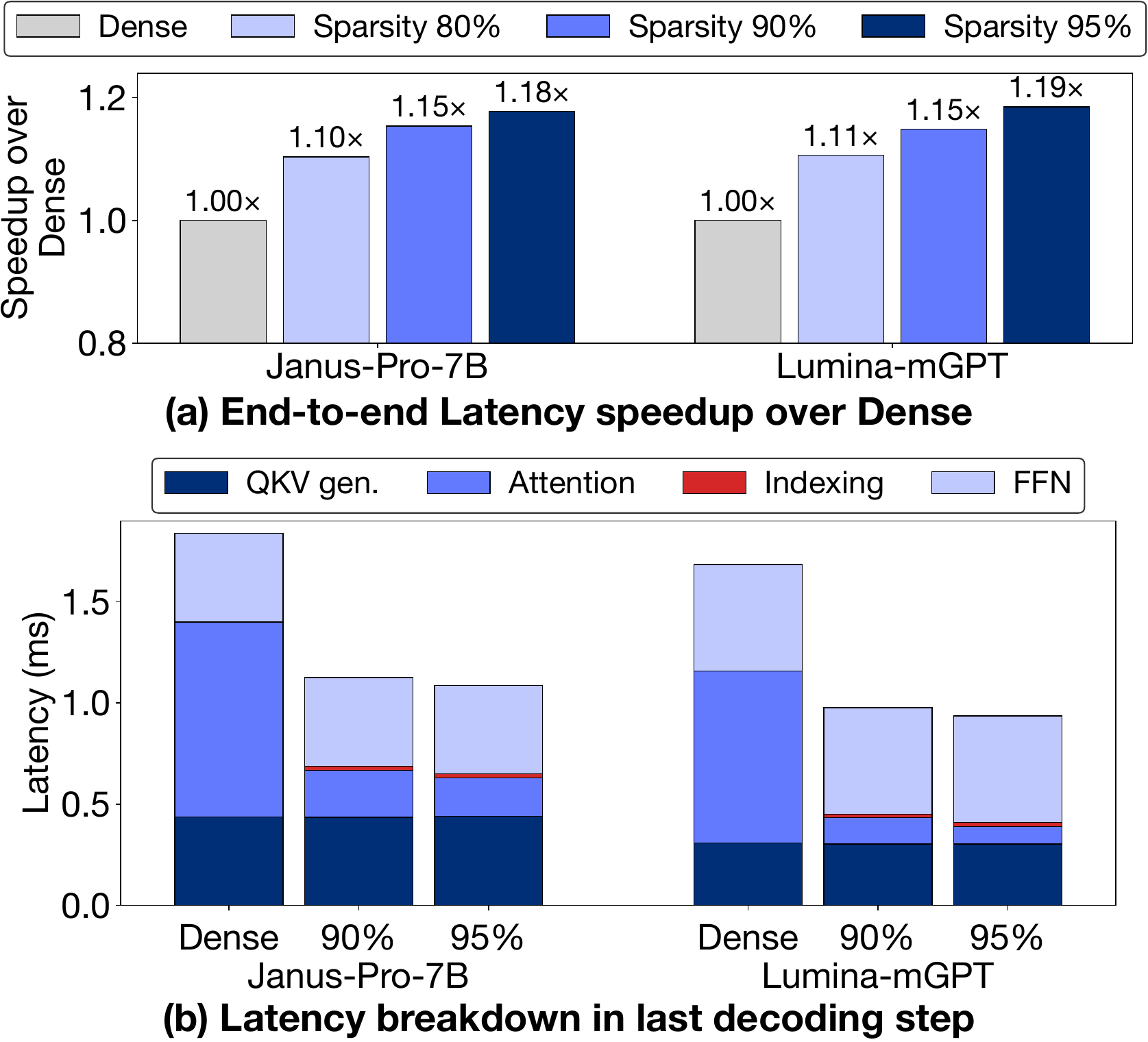}
\caption{(a) Latency speedup over dense and (b) latency breakdown in the last decoding step. Janus-Pro-7B used batch size 32 and Lumina-mGPT used batch size 16.}
\label{fig:latency-result}
\end{figure}

\begin{figure}[t]
\centering
\includegraphics[width=\linewidth]{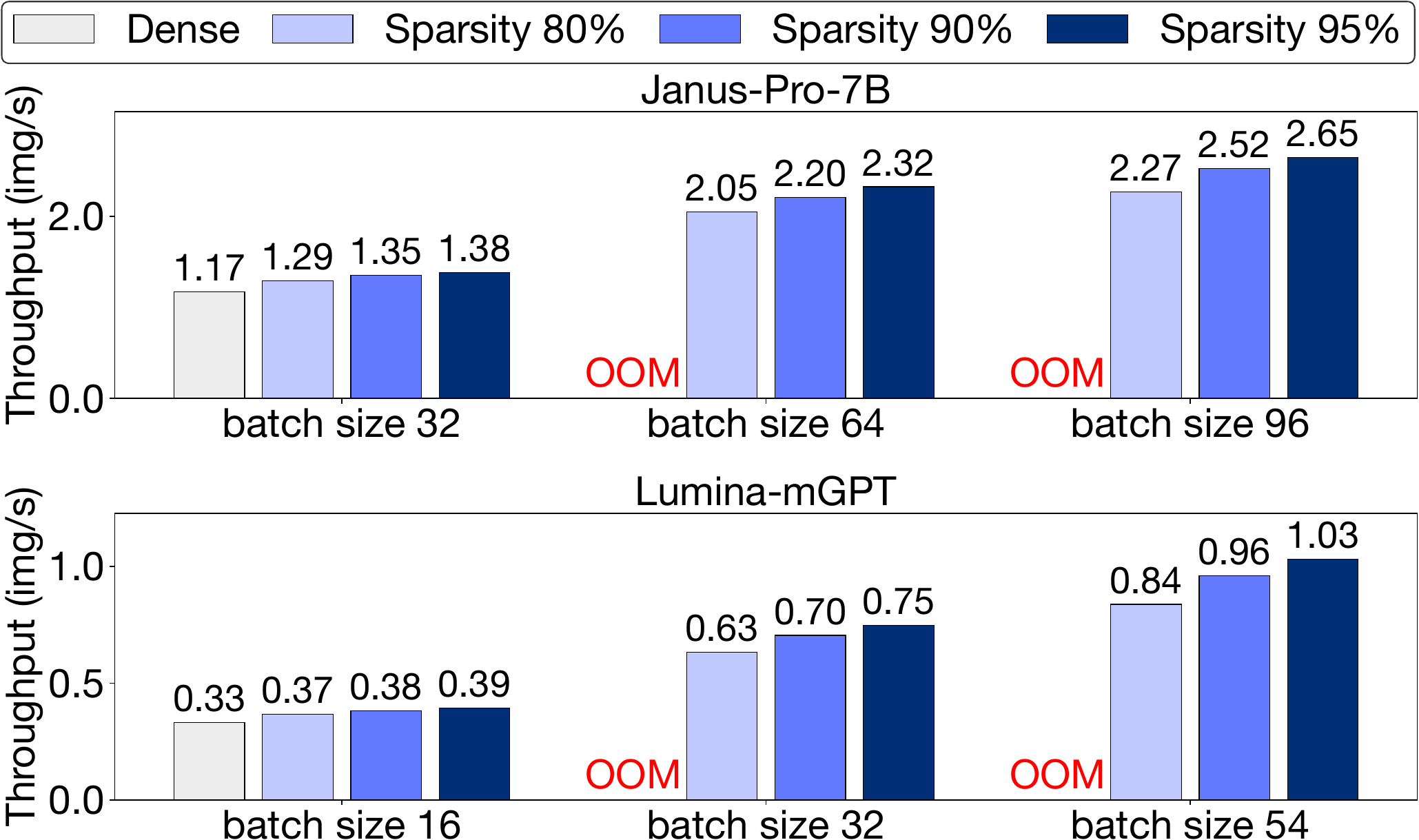}
\caption{Throughput improvement across different batch sizes on Janus-Pro-7B and Lumina-mGPT.}
\label{fig:throughput-result}
\vspace{-\baselineskip}
\end{figure}

\niparagraph{Tradeoff between latency and generation quality.}
Fig.~\ref{fig:latency-tradeoff} plots the quality-latency tradeoff for our method and StreamingLLM on Janus-Pro-7B and Lumina-mGPT. 
We compare against StreamingLLM as it achieves the highest quality among the baselines and incurs no selection overhead, making it the strongest reference point on both axes.
The x-axis shows end-to-end latency, and the y-axis shows the GenEval score. 
Moving up the plot reflects improved quality and moving left reflects reduced latency, so the top-left corner represents the most favorable tradeoff.
Diagonal dominates StreamingLLM across the sweep on both models. 
On Janus-Pro-7B at batch size 96, Diagonal+sink4 reaches dense-level GenEval score at 38s, while StreamingLLM requires 41s to reach comparable quality.
On Lumina-mGPT at batch size 54, Diagonal sustains 2-3\% of quality above StreamingLLM at every operating point, with the gap widening as latency tightens.
Diagonal selection recovers structurally important tokens that a fixed local window would otherwise drop, pushing the quality-latency frontier beyond the baseline.

\niparagraph{Latency speedup breakdown.}
Fig.~\ref{fig:latency-result}(a) shows the end-to-end latency speedup over dense model.
Speedup over dense reaches 1.18$\times$ and 1.19$\times$ at 95\% sparsity for Janus-Pro-7B and Lumina-mGPT, respectively. 
Fig.~\ref{fig:latency-result}(b) breaks down the per-step latency at the final decoding step into QKV projection, attention, sparse indexing overhead, and FFN.
At the final decoding step, attention alone speeds up by 4.15$\times$ and 5.03$\times$ over dense at 90\% and 95\% sparsity on Janus-Pro-7B, and by 6.52$\times$ and 9.78$\times$ on Lumina-mGPT. 
Indexing overhead stays small relative to the attention savings, as our custom kernel fuses indexing and attention  and removed redundant buffer allocation.
The attention latency shrinks with sparsity while indexing adds only a little overhead, leaving net latency well below dense.

\niparagraph{Throughput.}
Fig.~\ref{fig:throughput-result} reports per-second image throughput across batch sizes. 
First, the smaller KV footprint admits larger batches: dense Janus-Pro-7B runs out of memory at batch size 64 and beyond, while Diagonal scales to batch size 96. 
Lumina-mGPT shows the same pattern, with dense capped at batch size 16 while Diagonal reaches batch size 54. 
Second, within feasible batch sizes, per-image throughput improves with sparsity. 
On Janus-Pro-7B at batch size 32, Diagonal at 95\% sparsity reaches 1.38 img/s versus 1.17 img/s dense. 
At the largest feasible batch size, the combined effect yields up to 2.65 img/s on Janus-Pro-7B and 1.03 img/s on Lumina-mGPT, a 2.3$\times$ and 3.1$\times$ higher throughput over the dense configuration.
Overall, by keeping only the recent-window KV on the GPU, our method enables higher batch sizes, while its efficient sparse attention also raises throughput at a fixed batch size.

\section{Related Works}

\niparagraph{Image generation model.}
Diffusion models~\cite{nichol2021improved, rombach2022high, wu2025qwen, chen2024pixart} have enabled the practical deployment of image generation systems by overcoming major limitations of earlier GAN~\cite{goodfellow2014generative,radford2015unsupervised,karras2019style} based approaches, including low fidelity and training instability.
However, as discussed in Section~\ref{label:auto_image_generation}, diffusion backbones such as U-Net~\cite{ho2020denoising,song2020denoising,podell2024sdxl} and DiT~\cite{yang2208your,bao2023all,peebles2023scalable} remain architecturally distinct from LLMs, forcing multimodal serving frameworks to provision diffusion and LLM workloads separately, often partitioning GPU resources across different model pipelines~\cite{qiu2025modserve, yin2026vllm}.
In this respect, AR image generation~\cite{sun2024llamagen,wu2024janus,hu2024ella,chen2025janus,ma2024janusflow,liu2024lumina,wang2024emu3,team2025nextstep,ma2025token} offers greater compatibility with LLM-centric serving infrastructures.
Rather than replacing diffusion, AR image generation represents an orthogonal design point, and industry systems increasingly choose between diffusion and AR models based on serving-stack requirements~\cite{xai2024aurora,google2025gemini3proimagemodelcard,hurst2024gpt}.

\niparagraph{KV sparsity for LLM.}
The decoding phase of LLMs is memory-bandwidth-bound, and KV sparsity~\cite{zhang2023h2o,zhao2024alisa,xiao2023efficient,liu2023scissorhands,oren2024tova,lee2024infinigen} mitigates this by reducing the number of cache entries accessed at each step.
Beyond the basic position- and score-based schemes covered in Section~\ref{sec:bg-kv}, prior work has refined KV sparsity along several axes.
SnapKV~\cite{li2024snapkv} and NACL~\cite{chen2024nacl} move selection to the prefill stage to avoid per-step decisions, using a prompt-end observation window and proxy tokens.
PyramidKV~\cite{cai2024pyramidkv} and Ada-KV~\cite{feng2026ada} distribute cache capacity unevenly across layers and heads.
LESS~\cite{dong2024less} keeps a low-rank summary of dropped KV, DuoAttention~\cite{xiao2025duoattention} and CateKV~\cite{jiang2025catekv} splits heads into retrieval and streaming. 
Quest~\cite{tang2024quest} and OmniKV~\cite{hao2025omnikv} go further by preserving the full cache, reducing only per-step access via per-page key bounds and inter-layer selection reuse.
All of these designs target the attention patterns of text LLMs; our characterization in Section~\ref{sec:characterization} shows these patterns shift in autoregressive image generation, motivating the diagonal-aware policy in Section~\ref{sec:method}.

\niparagraph{KV quantization for LLM.}
KV quantization~\cite{kim2025oaken} addresses the same bandwidth bottleneck, lowering per-entry bit-width.
KIVI~\cite{liu2024kivi} and KVQuant~\cite{hooper2024kvquant} enable 2-bit KV caches through per-channel key and per-token value quantization, and Atom~\cite{zhao2024atom} and QServe~\cite{lin2025qserve} support low-bit KV serving through outlier-aware quantization.
KV quantization is orthogonal to our method and can be applied on top of it for further bandwidth savings.

\niparagraph{AR image generation optimizations.}
Image generation specific properties have been widely exploited to accelerate vision inference~\cite{lee2025tortoise,castillo2025adaptive,ma2024deepcache,kim2025mixdit}, and recent work extends this principle to AR image generation~\cite{he2024zipar,wang2026par,he2025neighboring, xiang2026var}.
ZipAR~\cite{he2024zipar}, PAR~\cite{wang2026par}, and NAR~\cite{he2025neighboring} all exploit the spatial locality of image tokens to decode multiple tokens in parallel, by overlapping or reordering the token sequences.
These approaches improve decoding parallelism, so they are orthogonal to our method and can be combined with it.
Closer to our work, HACK~\cite{qin2026head} and ADSA~\cite{xiang2025make} illustrate diagonal attention behavior of autoregressive image models.
HACK classifies heads by sparsity variance and falls back to score-based selection within a local window, while ADSA selects diverse middle tokens per step.
In contrast, we turn the diagonal into an explicit selection rule, and quantitatively characterize the pattern across multiple AR image generation models.

\section{Conclusion}
\noindent
This paper presents the first characterization of attention sparsity in autoregressive image generation and shows that its sparsity behaviors fundamentally differ from those established in text-based LLM inference. 
Our analysis reveals that autoregressive image generation exhibits unique structural properties, including diagonal attention sparsity arising from the spatial locality of visual tokens. 
Building on these observations, we demonstrate that exploiting workload-specific sparsity structures enables substantial throughput and latency improvements with minimal quality degradation. 
As multimodal AI systems continue evolving toward agentic and physically interactive workloads, efficient visual generation becomes an increasingly important systems challenge. 
We believe this work represents an initial step toward a broader direction of approximation-aware image generation serving, where practical efficiency is achieved not only through hardware scaling, but also through workload-specific compromises guided by the structural properties of visual generation workloads.

\section*{Acknowledgments}
\noindent
This work was partly supported by the IITP(Institute of Information \& Communications Technology Planning \& Evaluation)-ITRC(Information Technology Research Center) grant funded by the Korea government(MSIT)(IITP-2026-RS-2020-II201795), the IITP grant funded by the Korea government(MSIT) (No.RS-2024-00459797), and IITP under the Graduate School of Artificial Intelligence Semiconductor(IITP-2026-RS-2023-00256472) grant funded by the Korea government(MSIT).
The authors used Claude for code, text, and figures, but all outputs were reviewed carefully by the authors.


\bibliographystyle{IEEEtranS}
\bibliography{reference}

@article{chen2025janus,
  title={Janus-Pro: Unified Multimodal Understanding and Generation with Data and Model Scaling},
  author={Chen, Xiaokang and Wu, Zhiyu and Liu, Xingchao and Pan, Zizheng and Liu, Wen and Xie, Zhenda and Yu, Xingkai and Ruan, Chong},
  journal={arXiv preprint arXiv:2501.17811},
  year={2025}
}

@inproceedings{wu2024janus,
  title={Janus: Decoupling visual encoding for unified multimodal understanding and generation},
  author={Wu, Chengyue and Chen, Xiaokang and Wu, Zhiyu and Ma, Yiyang and Liu, Xingchao and Pan, Zizheng and Liu, Wen and Xie, Zhenda and Yu, Xingkai and Ruan, Chong and others},
  booktitle={Proceedings of the Computer Vision and Pattern Recognition Conference},
  pages={12966--12977},
  year={2025}
}

@inproceedings{ma2024janusflow,
  title={Janusflow: Harmonizing autoregression and rectified flow for unified multimodal understanding and generation},
  author={Ma, Yiyang and Liu, Xingchao and Chen, Xiaokang and Liu, Wen and Wu, Chengyue and Wu, Zhiyu and Pan, Zizheng and Xie, Zhenda and Zhang, Haowei and Yu, Xingkai and others},
  booktitle={Proceedings of the IEEE/CVF Conference on Computer Vision and Pattern Recognition},
  pages={7739--7751},
  year={2025}
}

@article{liu2024lumina,
  title={Lumina-mgpt: Illuminate flexible photorealistic text-to-image generation with multimodal generative pretraining},
  author={Liu, Dongyang and Zhao, Shitian and Zhuo, Le and Lin, Weifeng and Xin, Yi and Li, Xinyue and Qin, Qi and Qiao, Yu and Li, Hongsheng and Gao, Peng},
  journal={arXiv preprint arXiv:2408.02657},
  year={2024}
}

@article{wang2024emu3,
  title={Emu3: Next-token prediction is all you need},
  author={Wang, Xinlong and Zhang, Xiaosong and Luo, Zhengxiong and Sun, Quan and Cui, Yufeng and Wang, Jinsheng and Zhang, Fan and Wang, Yueze and Li, Zhen and Yu, Qiying and others},
  journal={arXiv preprint arXiv:2409.18869},
  year={2024}
}

@article{team2025nextstep,
  title={Nextstep-1: Toward autoregressive image generation with continuous tokens at scale},
  author={Team, NextStep and Han, Chunrui and Li, Guopeng and Wu, Jingwei and Sun, Quan and Cai, Yan and Peng, Yuang and Ge, Zheng and Zhou, Deyu and Tang, Haomiao and others},
  journal={arXiv preprint arXiv:2508.10711},
  year={2025}
}

@article{hu2024ella,
  title={Ella: Equip diffusion models with llm for enhanced semantic alignment},
  author={Hu, Xiwei and Wang, Rui and Fang, Yixiao and Fu, Bin and Cheng, Pei and Yu, Gang},
  journal={arXiv preprint arXiv:2403.05135},
  year={2024}
}

@article{ma2025token,
  title={Token-shuffle: Towards high-resolution image generation with autoregressive models},
  author={Ma, Xu and Sun, Peize and Ma, Haoyu and Tang, Hao and Ma, Chih-Yao and Wang, Jialiang and Li, Kunpeng and Dai, Xiaoliang and Shi, Yujun and Ju, Xuan and others},
  journal={arXiv preprint arXiv:2504.17789},
  year={2025}
}

@article{sun2024llamagen,
  title={Autoregressive Model Beats Diffusion: Llama for Scalable Image Generation},
  author={Sun, Peize and Jiang, Yi and Chen, Shoufa and Zhang, Shilong and Peng, Bingyue and Luo, Ping and Yuan, Zehuan},
  journal={arXiv preprint arXiv:2406.06525},
  year={2024}
}

@article{van2017neural,
  title={Neural discrete representation learning},
  author={Van Den Oord, Aaron and Vinyals, Oriol and Kavukcuoglu, Koray},
  journal={Advances in neural information processing systems},
  volume={30},
  year={2017}
}

@article{razavi2019generating,
  title={Generating diverse high-fidelity images with vq-vae-2},
  author={Razavi, Ali and Van den Oord, Aaron and Vinyals, Oriol},
  journal={Advances in neural information processing systems},
  volume={32},
  year={2019}
}

@misc{google2025gemini25flashmodelcard,
  author       = {Google},
  title        = {{2.5 Flash and Native Capabilities -- Audio \& Image: Model Card}},
  year         = {2025},
  month        = dec,
  note         = {Model card, published December 2025},
  url          = {https://storage.googleapis.com/deepmind-media/Model-Cards/Gemini-2-5-Flash-Model-Card.pdf}
}

@misc{google2025gemini3proimagemodelcard,
  author       = {Google},
  title        = {{Gemini 3 Pro Image: Model Card}},
  year         = {2025},
  month        = nov,
  note         = {Model card, published November 2025},
  url          = {https://storage.googleapis.com/deepmind-media/Model-Cards/Gemini-3-Pro-Image-Model-Card.pdf}
}

@article{hurst2024gpt,
  title={Gpt-4o system card},
  author={Hurst, Aaron and Lerer, Adam and Goucher, Adam P and Perelman, Adam and Ramesh, Aditya and Clark, Aidan and Ostrow, AJ and Welihinda, Akila and Hayes, Alan and Radford, Alec and others},
  journal={arXiv preprint arXiv:2410.21276},
  year={2024}
}

@misc{gemmateam2025gemma3technicalreport,
      title={Gemma 3 Technical Report}, 
      author={Gemma Team and Aishwarya Kamath and Johan Ferret and Shreya Pathak and Nino Vieillard and Ramona Merhej and Sarah Perrin and Tatiana Matejovicova and Alexandre Ramé and Morgane Rivière and Louis Rouillard and Thomas Mesnard and Geoffrey Cideron and Jean-bastien Grill and Sabela Ramos and Edouard Yvinec and Michelle Casbon and Etienne Pot and Ivo Penchev and Gaël Liu and Francesco Visin and Kathleen Kenealy and Lucas Beyer and Xiaohai Zhai and Anton Tsitsulin and Robert Busa-Fekete and Alex Feng and Noveen Sachdeva and Benjamin Coleman and Yi Gao and Basil Mustafa and Iain Barr and Emilio Parisotto and David Tian and Matan Eyal and Colin Cherry and Jan-Thorsten Peter and Danila Sinopalnikov and Surya Bhupatiraju and Rishabh Agarwal and Mehran Kazemi and Dan Malkin and Ravin Kumar and David Vilar and Idan Brusilovsky and Jiaming Luo and Andreas Steiner and Abe Friesen and Abhanshu Sharma and Abheesht Sharma and Adi Mayrav Gilady and Adrian Goedeckemeyer and Alaa Saade and Alex Feng and Alexander Kolesnikov and Alexei Bendebury and Alvin Abdagic and Amit Vadi and András György and André Susano Pinto and Anil Das and Ankur Bapna and Antoine Miech and Antoine Yang and Antonia Paterson and Ashish Shenoy and Ayan Chakrabarti and Bilal Piot and Bo Wu and Bobak Shahriari and Bryce Petrini and Charlie Chen and Charline Le Lan and Christopher A. Choquette-Choo and CJ Carey and Cormac Brick and Daniel Deutsch and Danielle Eisenbud and Dee Cattle and Derek Cheng and Dimitris Paparas and Divyashree Shivakumar Sreepathihalli and Doug Reid and Dustin Tran and Dustin Zelle and Eric Noland and Erwin Huizenga and Eugene Kharitonov and Frederick Liu and Gagik Amirkhanyan and Glenn Cameron and Hadi Hashemi and Hanna Klimczak-Plucińska and Harman Singh and Harsh Mehta and Harshal Tushar Lehri and Hussein Hazimeh and Ian Ballantyne and Idan Szpektor and Ivan Nardini and Jean Pouget-Abadie and Jetha Chan and Joe Stanton and John Wieting and Jonathan Lai and Jordi Orbay and Joseph Fernandez and Josh Newlan and Ju-yeong Ji and Jyotinder Singh and Kat Black and Kathy Yu and Kevin Hui and Kiran Vodrahalli and Klaus Greff and Linhai Qiu and Marcella Valentine and Marina Coelho and Marvin Ritter and Matt Hoffman and Matthew Watson and Mayank Chaturvedi and Michael Moynihan and Min Ma and Nabila Babar and Natasha Noy and Nathan Byrd and Nick Roy and Nikola Momchev and Nilay Chauhan and Noveen Sachdeva and Oskar Bunyan and Pankil Botarda and Paul Caron and Paul Kishan Rubenstein and Phil Culliton and Philipp Schmid and Pier Giuseppe Sessa and Pingmei Xu and Piotr Stanczyk and Pouya Tafti and Rakesh Shivanna and Renjie Wu and Renke Pan and Reza Rokni and Rob Willoughby and Rohith Vallu and Ryan Mullins and Sammy Jerome and Sara Smoot and Sertan Girgin and Shariq Iqbal and Shashir Reddy and Shruti Sheth and Siim Põder and Sijal Bhatnagar and Sindhu Raghuram Panyam and Sivan Eiger and Susan Zhang and Tianqi Liu and Trevor Yacovone and Tyler Liechty and Uday Kalra and Utku Evci and Vedant Misra and Vincent Roseberry and Vlad Feinberg and Vlad Kolesnikov and Woohyun Han and Woosuk Kwon and Xi Chen and Yinlam Chow and Yuvein Zhu and Zichuan Wei and Zoltan Egyed and Victor Cotruta and Minh Giang and Phoebe Kirk and Anand Rao and Kat Black and Nabila Babar and Jessica Lo and Erica Moreira and Luiz Gustavo Martins and Omar Sanseviero and Lucas Gonzalez and Zach Gleicher and Tris Warkentin and Vahab Mirrokni and Evan Senter and Eli Collins and Joelle Barral and Zoubin Ghahramani and Raia Hadsell and Yossi Matias and D. Sculley and Slav Petrov and Noah Fiedel and Noam Shazeer and Oriol Vinyals and Jeff Dean and Demis Hassabis and Koray Kavukcuoglu and Clement Farabet and Elena Buchatskaya and Jean-Baptiste Alayrac and Rohan Anil and Dmitry and Lepikhin and Sebastian Borgeaud and Olivier Bachem and Armand Joulin and Alek Andreev and Cassidy Hardin and Robert Dadashi and Léonard Hussenot},
      year={2025},
      eprint={2503.19786},
      archivePrefix={arXiv},
      primaryClass={cs.CL},
      url={https://arxiv.org/abs/2503.19786}, 
}

@article{yang2025qwen3,
  title={Qwen3 technical report},
  author={Yang, An and Li, Anfeng and Yang, Baosong and Zhang, Beichen and Hui, Binyuan and Zheng, Bo and Yu, Bowen and Gao, Chang and Huang, Chengen and Lv, Chenxu and others},
  journal={arXiv preprint arXiv:2505.09388},
  year={2025}
}

@article{grattafiori2024llama,
  title={The llama 3 herd of models},
  author={Grattafiori, Aaron and Dubey, Abhimanyu and Jauhri, Abhinav and Pandey, Abhinav and Kadian, Abhishek and Al-Dahle, Ahmad and Letman, Aiesha and Mathur, Akhil and Schelten, Alan and Vaughan, Alex and others},
  journal={arXiv preprint arXiv:2407.21783},
  year={2024}
}

@article{zaheer2020big,
  title={Big bird: Transformers for longer sequences},
  author={Zaheer, Manzil and Guruganesh, Guru and Dubey, Kumar Avinava and Ainslie, Joshua and Alberti, Chris and Ontanon, Santiago and Pham, Philip and Ravula, Anirudh and Wang, Qifan and Yang, Li and others},
  journal={Advances in neural information processing systems},
  volume={33},
  pages={17283--17297},
  year={2020}
}

@article{beltagy2020longformer,
  title={Longformer: The long-document transformer},
  author={Beltagy, Iz and Peters, Matthew E and Cohan, Arman},
  journal={arXiv preprint arXiv:2004.05150},
  year={2020}
}

@article{zhang2023h2o,
  title={H2o: Heavy-hitter oracle for efficient generative inference of large language models},
  author={Zhang, Zhenyu and Sheng, Ying and Zhou, Tianyi and Chen, Tianlong and Zheng, Lianmin and Cai, Ruisi and Song, Zhao and Tian, Yuandong and R{\'e}, Christopher and Barrett, Clark and others},
  journal={Advances in Neural Information Processing Systems},
  volume={36},
  pages={34661--34710},
  year={2023}
}

@inproceedings{zhao2024alisa,
  title={Alisa: Accelerating large language model inference via sparsity-aware kv caching},
  author={Zhao, Youpeng and Wu, Di and Wang, Jun},
  booktitle={2024 ACM/IEEE 51st Annual International Symposium on Computer Architecture (ISCA)},
  pages={1005--1017},
  year={2024},
  organization={IEEE}
}

@inproceedings{xiao2023efficient,
  title={Efficient streaming language models with attention sinks},
  author={Xiao, Guangxuan and Tian, Yuandong and Chen, Beidi and Han, Song and Lewis, Mike},
  booktitle={International Conference on Learning Representations},
  volume={2024},
  pages={21875--21895},
  year={2024}
}

@inproceedings{lee2024infinigen,
  title={$\{$InfiniGen$\}$: Efficient generative inference of large language models with dynamic $\{$KV$\}$ cache management},
  author={Lee, Wonbeom and Lee, Jungi and Seo, Junghwan and Sim, Jaewoong},
  booktitle={18th USENIX Symposium on Operating Systems Design and Implementation (OSDI 24)},
  pages={155--172},
  year={2024}
}

@inproceedings{oren2024tova,
  title={Transformers are multi-state rnns},
  author={Oren, Matanel and Hassid, Michael and Yarden, Nir and Adi, Yossi and Schwartz, Roy},
  booktitle={Proceedings of the 2024 Conference on Empirical Methods in Natural Language Processing},
  pages={18724--18741},
  year={2024}
}

@article{tang2024quest,
  title={Quest: Query-aware sparsity for efficient long-context llm inference},
  author={Tang, Jiaming and Zhao, Yilong and Zhu, Kan and Xiao, Guangxuan and Kasikci, Baris and Han, Song},
  journal={arXiv preprint arXiv:2406.10774},
  year={2024}
}

@inproceedings{xiao2025duoattention,
  title={Duoattention: Efficient long-context llm inference with retrieval and streaming heads},
  author={Xiao, Guangxuan and Tang, Jiaming and Zuo, Jingwei and Guo, Junxian and Yang, Shang and Tang, Haotian and Fu, Yao and Han, Song},
  booktitle={International Conference on Learning Representations},
  volume={2025},
  pages={37228--37253},
  year={2025}
}

@inproceedings{hao2025omnikv,
  title={Omnikv: Dynamic context selection for efficient long-context llms},
  author={Hao, Jitai and Zhu, Yuke and Wang, Tian and Yu, Jun and Xin, Xin and Zheng, Bo and Ren, Zhaochun and Guo, Sheng},
  booktitle={The Thirteenth International Conference on Learning Representations},
  year={2025}
}

@article{liu2023scissorhands,
  title={Scissorhands: Exploiting the persistence of importance hypothesis for llm kv cache compression at test time},
  author={Liu, Zichang and Desai, Aditya and Liao, Fangshuo and Wang, Weitao and Xie, Victor and Xu, Zhaozhuo and Kyrillidis, Anastasios and Shrivastava, Anshumali},
  journal={Advances in Neural Information Processing Systems},
  volume={36},
  pages={52342--52364},
  year={2023}
}

@inproceedings{chen2024nacl,
  title={Nacl: A general and effective kv cache eviction framework for llm at inference time},
  author={Chen, Yilong and Wang, Guoxia and Shang, Junyuan and Cui, Shiyao and Zhang, Zhenyu and Liu, Tingwen and Wang, Shuohuan and Sun, Yu and Yu, Dianhai and Wu, Hua},
  booktitle={Proceedings of the 62nd Annual Meeting of the Association for Computational Linguistics (Volume 1: Long Papers)},
  pages={7913--7926},
  year={2024}
}

@article{li2024snapkv,
  title={Snapkv: Llm knows what you are looking for before generation},
  author={Li, Yuhong and Huang, Yingbing and Yang, Bowen and Venkitesh, Bharat and Locatelli, Acyr and Ye, Hanchen and Cai, Tianle and Lewis, Patrick and Chen, Deming},
  journal={Advances in Neural Information Processing Systems},
  volume={37},
  pages={22947--22970},
  year={2024}
}

@article{feng2026ada,
  title={Ada-kv: Optimizing kv cache eviction by adaptive budget allocation for efficient llm inference},
  author={Feng, Yuan and Lv, Junlin and Cao, Yukun and Xie, Xike and Zhou, S Kevin},
  journal={Advances in Neural Information Processing Systems},
  volume={38},
  pages={113152--113188},
  year={2026}
}

@article{dong2024less,
  title={Get more with less: Synthesizing recurrence with kv cache compression for efficient llm inference},
  author={Dong, Harry and Yang, Xinyu and Zhang, Zhenyu and Wang, Zhangyang and Chi, Yuejie and Chen, Beidi},
  journal={arXiv preprint arXiv:2402.09398},
  year={2024}
}

@article{cai2024pyramidkv,
  title={Pyramidkv: Dynamic kv cache compression based on pyramidal information funneling},
  author={Cai, Zefan and Zhang, Yichi and Gao, Bofei and Liu, Yuliang and Li, Yucheng and Liu, Tianyu and Lu, Keming and Xiong, Wayne and Dong, Yue and Hu, Junjie and others},
  journal={arXiv preprint arXiv:2406.02069},
  year={2024}
}

@article{he2024zipar,
  title={ZipAR: Parallel Auto-regressive Image Generation through Spatial Locality},
  author={He, Yefei and Chen, Feng and He, Yuanyu and He, Shaoxuan and Zhou, Hong and Zhang, Kaipeng and Zhuang, Bohan},
  journal={arXiv preprint arXiv:2412.04062},
  year={2024}
}

@article{wang2026par,
  title={Conditional panoramic image generation via masked autoregressive modeling},
  author={Wang, Chaoyang and Li, Xiangtai and Qi, Lu and Lin, Xiaofan and Bai, Jinbin and Zhou, Qianyu and Tong, Yunhai},
  journal={Advances in Neural Information Processing Systems},
  volume={38},
  pages={27654--27679},
  year={2026}
}

@inproceedings{he2025neighboring,
  title={Neighboring autoregressive modeling for efficient visual generation},
  author={He, Yefei and He, Yuanyu and He, Shaoxuan and Chen, Feng and Zhou, Hong and Zhang, Kaipeng and Zhuang, Bohan},
  booktitle={Proceedings of the IEEE/CVF International Conference on Computer Vision},
  pages={19000--19010},
  year={2025}
}

@article{lee2025tortoise,
  title={Tortoise and Hare Guidance: Accelerating Diffusion Model Inference with Multirate Integration},
  author={Lee, Yunghee and Pak, Byeonghyun and Hong, Junwha and Kim, Hoseong},
  journal={Advances in Neural Information Processing Systems},
  volume={38},
  pages={2871--2898},
  year={2026}
}

@inproceedings{castillo2025adaptive,
  title={Adaptive guidance: Training-free acceleration of conditional diffusion models},
  author={Castillo, Angela and Kohler, Jonas and P{\'e}rez, Juan C and P{\'e}rez, Juan Pablo and Pumarola, Albert and Ghanem, Bernard and Arbel{\'a}ez, Pablo and Thabet, Ali},
  booktitle={Proceedings of the AAAI Conference on Artificial Intelligence},
  volume={39},
  number={2},
  pages={1962--1970},
  year={2025}
}

@article{ghosh2023geneval,
  title={Geneval: An object-focused framework for evaluating text-to-image alignment},
  author={Ghosh, Dhruba and Hajishirzi, Hannaneh and Schmidt, Ludwig},
  journal={Advances in Neural Information Processing Systems},
  volume={36},
  pages={52132--52152},
  year={2023}
}

@article{sun2023journeydb,
  title={Journeydb: A benchmark for generative image understanding},
  author={Sun, Keqiang and Pan, Junting and Ge, Yuying and Li, Hao and Duan, Haodong and Wu, Xiaoshi and Zhang, Renrui and Zhou, Aojun and Qin, Zipeng and Wang, Yi and others},
  journal={Advances in neural information processing systems},
  volume={36},
  pages={49659--49678},
  year={2023}
}

@misc{civitprompts,
  author       = {{AdamCodd}},
  title        = {Civitai 2M prompts},
  year         = {2024},
  publisher    = {Hugging Face},
  howpublished = {\url{https://huggingface.co/datasets/AdamCodd/Civitai-2m-prompts}},
  note         = {Accessed: 2026-04-27}
}

@inproceedings{wang2023diffusiondb,
  title={Diffusiondb: A large-scale prompt gallery dataset for text-to-image generative models},
  author={Wang, Zijie J and Montoya, Evan and Munechika, David and Yang, Haoyang and Hoover, Benjamin and Chau, Duen Horng},
  booktitle={Proceedings of the 61st annual meeting of the association for computational linguistics (volume 1: Long papers)},
  pages={893--911},
  year={2023}
}

@misc{azure_llm_inference,
  author       = {{Microsoft Azure}},
  title        = {{Azure LLM Inference Trace 2025}},
  year         = {2025},
  publisher    = {GitHub},
  howpublished = {\url{https://github.com/Azure/AzurePublicDataset}},
  note         = {Accessed: 2026-04-22}
}

@inproceedings{sheng2023flexgen,
  title={Flexgen: High-throughput generative inference of large language models with a single gpu},
  author={Sheng, Ying and Zheng, Lianmin and Yuan, Binhang and Li, Zhuohan and Ryabinin, Max and Chen, Beidi and Liang, Percy and R{\'e}, Christopher and Stoica, Ion and Zhang, Ce},
  booktitle={International Conference on Machine Learning},
  pages={31094--31116},
  year={2023},
  organization={PMLR}
}

@inproceedings{dao2024flashattention,
  title={Flashattention-2: Faster attention with better parallelism and work partitioning},
  author={Dao, Tri},
  booktitle={International Conference on Learning Representations},
  volume={2024},
  pages={35549--35562},
  year={2024}
}

@inproceedings{qiu2025modserve,
  title={Modserve: Modality-and stage-aware resource disaggregation for scalable multimodal model serving},
  author={Qiu, Haoran and Biswas, Anish and Zhao, Zihan and Mohan, Jayashree and Khare, Alind and Choukse, Esha and Goiri, {\'I}{\~n}igo and Zhang, Zeyu and Shen, Haiying and Bansal, Chetan and Ramjee, Ramachandran and Fonseca, Rodrigo},
  booktitle={Proceedings of the 2025 ACM Symposium on Cloud Computing},
  pages={817--830},
  year={2025}
}

@article{yin2026vllm,
  title={vLLM-Omni: Fully Disaggregated Serving for Any-to-Any Multimodal Models},
  author={Yin, Peiqi and Zhu, Jiangyun and Gao, Han and Zheng, Chenguang and Huang, Yongxiang and Zhou, Taichang and Yang, Ruirui and Liu, Weizhi and Chen, Weiqing and Guo, Canlin and others},
  journal={arXiv preprint arXiv:2602.02204},
  year={2026}
}

@article{liu2024kivi,
  title={Kivi: A tuning-free asymmetric 2bit quantization for kv cache},
  author={Liu, Zirui and Yuan, Jiayi and Jin, Hongye and Zhong, Shaochen and Xu, Zhaozhuo and Braverman, Vladimir and Chen, Beidi and Hu, Xia},
  journal={arXiv preprint arXiv:2402.02750},
  year={2024}
}

@article{hooper2024kvquant,
  title={Kvquant: Towards 10 million context length llm inference with kv cache quantization},
  author={Hooper, Coleman and Kim, Sehoon and Mohammadzadeh, Hiva and Mahoney, Michael W and Shao, Yakun S and Keutzer, Kurt and Gholami, Amir},
  journal={Advances in Neural Information Processing Systems},
  volume={37},
  pages={1270--1303},
  year={2024}
}

@article{zhao2024atom,
  title={Atom: Low-bit quantization for efficient and accurate llm serving},
  author={Zhao, Yilong and Lin, Chien-Yu and Zhu, Kan and Ye, Zihao and Chen, Lequn and Zheng, Size and Ceze, Luis and Krishnamurthy, Arvind and Chen, Tianqi and Kasikci, Baris},
  journal={Proceedings of Machine Learning and Systems},
  volume={6},
  pages={196--209},
  year={2024}
}

@article{lin2025qserve,
  title={Qserve: W4a8kv4 quantization and system co-design for efficient llm serving},
  author={Lin, Yujun and Tang, Haotian and Yang, Shang and Zhang, Zhekai and Xiao, Guangxuan and Gan, Chuang and Han, Song},
  journal={Proceedings of Machine Learning and Systems},
  volume={7},
  year={2025}
}

@article{goodfellow2014generative,
  title={Generative adversarial nets},
  author={Goodfellow, Ian J and Pouget-Abadie, Jean and Mirza, Mehdi and Xu, Bing and Warde-Farley, David and Ozair, Sherjil and Courville, Aaron and Bengio, Yoshua},
  journal={Advances in neural information processing systems},
  volume={27},
  year={2014}
}

@article{radford2015unsupervised,
  title={Unsupervised representation learning with deep convolutional generative adversarial networks},
  author={Radford, Alec and Metz, Luke and Chintala, Soumith},
  journal={arXiv preprint arXiv:1511.06434},
  year={2015}
}

@inproceedings{karras2019style,
  title={A style-based generator architecture for generative adversarial networks},
  author={Karras, Tero and Laine, Samuli and Aila, Timo},
  booktitle={Proceedings of the IEEE/CVF conference on computer vision and pattern recognition},
  pages={4401--4410},
  year={2019}
}

@article{ho2020denoising,
  title={Denoising diffusion probabilistic models},
  author={Ho, Jonathan and Jain, Ajay and Abbeel, Pieter},
  journal={Advances in neural information processing systems},
  volume={33},
  pages={6840--6851},
  year={2020}
}

@article{song2020denoising,
  title={Denoising diffusion implicit models},
  author={Song, Jiaming and Meng, Chenlin and Ermon, Stefano},
  journal={arXiv preprint arXiv:2010.02502},
  year={2020}
}

@inproceedings{nichol2021improved,
  title={Improved denoising diffusion probabilistic models},
  author={Nichol, Alexander Quinn and Dhariwal, Prafulla},
  booktitle={International conference on machine learning},
  pages={8162--8171},
  year={2021},
  organization={PMLR}
}

@inproceedings{rombach2022high,
  title={High-resolution image synthesis with latent diffusion models},
  author={Rombach, Robin and Blattmann, Andreas and Lorenz, Dominik and Esser, Patrick and Ommer, Bj{\"o}rn},
  booktitle={Proceedings of the IEEE/CVF conference on computer vision and pattern recognition},
  pages={10684--10695},
  year={2022}
}

@article{yang2208your,
  title={Your vit is secretly a hybrid discriminative-generative diffusion model. arXiv 2022},
  author={Yang, X and Shih, SM and Fu, Y and Zhao, X and Ji, S},
  journal={arXiv preprint arXiv:2208.07791}
}

@inproceedings{bao2023all,
  title={All are worth words: A vit backbone for diffusion models},
  author={Bao, Fan and Nie, Shen and Xue, Kaiwen and Cao, Yue and Li, Chongxuan and Su, Hang and Zhu, Jun},
  booktitle={Proceedings of the IEEE/CVF conference on computer vision and pattern recognition},
  pages={22669--22679},
  year={2023}
}

@article{hwang2025d,
  title={D$\backslash$'ej$\backslash$a Vu: Efficient Video-Language Query Engine with Learning-based Inter-Frame Computation Reuse},
  author={Hwang, Jinwoo and Kim, Daeun and Lee, Sangyeop and Kim, Yoonsung and Heo, Guseul and Kim, Hojoon and Jeong, Yunseok and Meaza, Tadiwos and Park, Eunhyeok and Ahn, Jeongseob and others},
  journal={arXiv preprint arXiv:2506.14107},
  year={2025}
}

@inproceedings{oh2026neo,
  title={Neo: Real-Time On-Device 3D Gaussian Splatting with Reuse-and-Update Sorting Acceleration},
  author={Oh, Changhun and Oh, Seongryong and Hwang, Jinwoo and Kim, Yoonsung and Sharma, Hardik and Park, Jongse},
  booktitle={Proceedings of the 31st ACM International Conference on Architectural Support for Programming Languages and Operating Systems, Volume 2},
  pages={1268--1284},
  year={2026}
}

@inproceedings{peebles2023scalable,
  title={Scalable diffusion models with transformers},
  author={Peebles, William and Xie, Saining},
  booktitle={Proceedings of the IEEE/CVF international conference on computer vision},
  pages={4195--4205},
  year={2023}
}

@inproceedings{tillet2019triton,
  title={Triton: an intermediate language and compiler for tiled neural network computations},
  author={Tillet, Philippe and Kung, Hsiang-Tsung and Cox, David},
  booktitle={Proceedings of the 3rd ACM SIGPLAN International Workshop on Machine Learning and Programming Languages},
  pages={10--19},
  year={2019}
}

@online{midjourney_draft,
  author  = {Midjourney},
  title   = {Draft and Conversational Modes},
  year    = {2025},
  url     = {https://docs.midjourney.com/hc/en-us/articles/35577175650957-Draft-Conversational-Modes},
  note = {Accessed: 2026-05-20}
}

@online{adobe_firefly_fastmode,
  author = {{Adobe}},
  title  = {Generate Images Quickly Using {F}ast Mode --- {A}dobe {F}irefly {W}eb {H}elp},
  year   = {2025},
  url    = {https://helpx.adobe.com/firefly/web/generate-images-with-text-to-image/generate-images-using-text-prompts/use-fast-mode-for-quick-image-generations.html},
  note = {Accessed: 2026-05-20}
}

@online{sd_safety_checker,
  title={{S}table {D}iffusion Safety Checker},
  author={{CompVis} and {Stability AI}},
  year={2022},
  url={https://huggingface.co/CompVis/stable-diffusion-safety-checker},
  note={Accessed: 2026-05-22}
}

@inproceedings{liu2026wukong,
  title={Wukong framework for not safe for work detection in text-to-image systems},
  author={Liu, Mingrui and Zhang, Sixiao and Long, Cheng},
  booktitle={Proceedings of the 32nd ACM SIGKDD Conference on Knowledge Discovery and Data Mining V. 1},
  pages={891--902},
  year={2026}
}

@article{hartmann2025genmarketing,
  author  = {Hartmann, Jochen and Exner, Yannick and Domdey, Samuel},
  title   = {The Power of Generative Marketing: 
             Can Generative {AI} Create Superhuman Visual 
             Marketing Content?},
  journal = {International Journal of Research in Marketing},
  volume  = {42},
  number  = {1},
  pages   = {13--31},
  year    = {2025},
  publisher = {Elsevier}
}

@article{betker2023improving,
  title={Improving image generation with better captions},
  author={Betker, James and Goh, Gabriel and Jing, Li and Brooks, Tim and Wang, Jianfeng and Li, Linjie and Ouyang, Long and Zhuang, Juntang and Lee, Joyce and Guo, Yufei and others},
  journal={Computer Science. https://cdn. openai. com/papers/dall-e-3. pdf},
  volume={2},
  number={3},
  pages={8},
  year={2023}
}

@online{midjourney,
  author = {Midjourney},
  title = {Midjourney},
  year = {2022},
  howpublished = {\url{https://www.midjourney.com}},
  note = {Accessed: 2026-05-20}
}

@online{xai2024aurora,
  author       = {{xAI}},
  title        = {Grok Image Generation Release},
  year         = {2024},
  url          = {https://x.ai/news/grok-image-generation-release},
  note         = {Accessed: 2026-05-20}
}

@online{gpt4osystemcard,
  author       = {{OpenAI}},
  title        = {Addendum to GPT-4o System Card: Native image generation},
  year         = {2025},
  url          = {https://cdn.openai.com/11998be9-5319-4302-bfbf-1167e093f1fb/Native_Image_Generation_System_Card.pdf},
  note         = {Accessed: 2026-05-20}
}

@inproceedings{chen2024pixart,
  title={Pixart-alpha: Fast training of diffusion transformer for photorealistic text-to-image synthesis},
  author={Chen, Junsong and Yu, Jincheng and Ge, Chongjian and Yao, Lewei and Xie, Enze and Wang, Zhongdao and Kwok, James and Luo, Ping and Lu, Huchuan and Li, Zhenguo},
  booktitle={International conference on learning representations},
  volume={2024},
  pages={57611--57640},
  year={2024}
}

@article{wu2025qwen,
  title={Qwen-image technical report},
  author={Wu, Chenfei and Li, Jiahao and Zhou, Jingren and Lin, Junyang and Gao, Kaiyuan and Yan, Kun and Yin, Sheng-ming and Bai, Shuai and Xu, Xiao and Chen, Yilei and others},
  journal={arXiv preprint arXiv:2508.02324},
  year={2025}
}

@inproceedings{ma2024deepcache,
  title={Deepcache: Accelerating diffusion models for free},
  author={Ma, Xinyin and Fang, Gongfan and Wang, Xinchao},
  booktitle={Proceedings of the IEEE/CVF conference on computer vision and pattern recognition},
  pages={15762--15772},
  year={2024}
}

@inproceedings{xiang2026var,
  title={VAR-Turbo: Unlocking the Potential of Visual Autoregressive Models Through Dual Redundancy},
  author={Xiang, Xujiang and Tu, Fengbin},
  booktitle={2026 IEEE International Symposium on High Performance Computer Architecture (HPCA)},
  pages={1--16},
  year={2026},
  organization={IEEE}
}

@inproceedings{jiang2025catekv,
  title={CateKV: On Sequential Consistency for Long-Context LLM Inference Acceleration},
  author={Jiang, Haoyun and Huang, Fei and Hu, Qiang and Sun, Minmin and Xiao, Shuai and Li, Yong and Lin, Junyang and Yao, Jiangchao and others},
  booktitle={Forty-second International Conference on Machine Learning},
  year={2025}
}

@inproceedings{podell2024sdxl,
  title={Sdxl: Improving latent diffusion models for high-resolution image synthesis},
  author={Podell, Dustin and English, Zion and Lacey, Kyle and Blattmann, Andreas and Dockhorn, Tim and M{\"u}ller, Jonas and Penna, Joe and Rombach, Robin},
  booktitle={International Conference on Learning Representations},
  volume={2024},
  pages={1862--1874},
  year={2024}
}

@article{yao2025survey,
  title={A survey on agentic multimodal large language models},
  author={Yao, Huanjin and Zhang, Ruifei and Huang, Jiaxing and Zhang, Jingyi and Wang, Yibo and Fang, Bo and Zhu, Ruolin and Jing, Yongcheng and Liu, Shunyu and Li, Guanbin and others},
  journal={arXiv preprint arXiv:2510.10991},
  year={2025}
}

@article{agarwal2025cosmos,
  title={Cosmos world foundation model platform for physical ai},
  author={Agarwal, Niket and Ali, Arslan and Bala, Maciej and Balaji, Yogesh and Barker, Erik and Cai, Tiffany and Chattopadhyay, Prithvijit and Chen, Yongxin and Cui, Yin and Ding, Yifan and others},
  journal={arXiv preprint arXiv:2501.03575},
  year={2025}
}

@article{han2015deep,
  title={Deep compression: Compressing deep neural networks with pruning, trained quantization and huffman coding},
  author={Han, Song and Mao, Huizi and Dally, William J},
  journal={arXiv preprint arXiv:1510.00149},
  year={2015}
}

@inproceedings{ham20203,
  title={A\^{} 3: Accelerating attention mechanisms in neural networks with approximation},
  author={Ham, Tae Jun and Jung, Sung Jun and Kim, Seonghak and Oh, Young H and Park, Yeonhong and Song, Yoonho and Park, Jung-Hun and Lee, Sanghee and Park, Kyoung and Lee, Jae W and others},
  booktitle={2020 IEEE International Symposium on High Performance Computer Architecture (HPCA)},
  pages={328--341},
  year={2020},
  organization={IEEE}
}

@inproceedings{wang2021spatten,
  title={Spatten: Efficient sparse attention architecture with cascade token and head pruning},
  author={Wang, Hanrui and Zhang, Zhekai and Han, Song},
  booktitle={2021 IEEE international symposium on high-performance computer architecture (HPCA)},
  pages={97--110},
  year={2021},
  organization={IEEE}
}

@inproceedings{you2023vitcod,
  title={Vitcod: Vision transformer acceleration via dedicated algorithm and accelerator co-design},
  author={You, Haoran and Sun, Zhanyi and Shi, Huihong and Yu, Zhongzhi and Zhao, Yang and Zhang, Yongan and Li, Chaojian and Li, Baopu and Lin, Yingyan},
  booktitle={2023 IEEE International Symposium on High-Performance Computer Architecture (HPCA)},
  pages={273--286},
  year={2023},
  organization={IEEE}
}

@inproceedings{hwang2022cova,
  title={$\{$CoVA$\}$: Exploiting $\{$Compressed-Domain$\}$ analysis to accelerate video analytics},
  author={Hwang, Jinwoo and Kim, Minsu and Kim, Daeun and Nam, Seungho and Kim, Yoonsung and Kim, Dohee and Sharma, Hardik and Park, Jongse},
  booktitle={2022 USENIX annual technical conference (USENIX ATC 22)},
  pages={707--722},
  year={2022}
}

@inproceedings{park2016axgames,
author = {Park, Jongse and Amaro, Emmanuel and Mahajan, Divya and Thwaites, Bradley and Esmaeilzadeh, Hadi},
title = {AxGames: Towards Crowdsourcing Quality Target Determination in Approximate Computing},
year = {2016},
publisher = {Association for Computing Machinery},
address = {New York, NY, USA},
doi = {10.1145/2872362.2872376},
booktitle = {Proceedings of the Twenty-First International Conference on Architectural Support for Programming Languages and Operating Systems},
pages = {623–636},
series = {ASPLOS '16}
}

@article{guenter2012foveated,
  title={Foveated 3D graphics},
  author={Guenter, Brian and Finch, Mark and Drucker, Steven and Tan, Desney and Snyder, John},
  journal={ACM transactions on Graphics (tOG)},
  volume={31},
  number={6},
  pages={1--10},
  year={2012},
  publisher={ACM New York, NY, USA}
}

@article{patney2016towards,
  title={Towards foveated rendering for gaze-tracked virtual reality},
  author={Patney, Anjul and Salvi, Marco and Kim, Joohwan and Kaplanyan, Anton and Wyman, Chris and Benty, Nir and Luebke, David and Lefohn, Aaron},
  journal={ACM Transactions On Graphics (TOG)},
  volume={35},
  number={6},
  pages={1--12},
  year={2016},
  publisher={ACM New York, NY, USA}
}

@article{wang2023foveated,
  title={Foveated rendering: A state-of-the-art survey},
  author={Wang, Lili and Shi, Xuehuai and Liu, Yi},
  journal={Computational visual media},
  volume={9},
  number={2},
  pages={195--228},
  year={2023},
  publisher={TUP}
}

@article{kim2025mixdit,
  title={Mixdit: Accelerating image diffusion transformer inference with mixed-precision mx quantization},
  author={Kim, Daeun and Hwang, Jinwoo and Oh, Changhun and Park, Jongse},
  journal={IEEE Computer Architecture Letters},
  volume={24},
  number={1},
  pages={141--144},
  year={2025},
  publisher={IEEE}
}

@inproceedings{kim2025oaken,
  title={Oaken: Fast and efficient llm serving with online-offline hybrid kv cache quantization},
  author={Kim, Minsu and Hong, Seongmin and Ko, RyeoWook and Choi, Soongyu and Lee, Hunjong and Kim, Junsoo and Kim, Joo-Young and Park, Jongse},
  booktitle={Proceedings of the 52nd Annual International Symposium on Computer Architecture},
  pages={482--497},
  year={2025}
}

@inproceedings{qin2026head,
  title={Head-aware kv cache compression for efficient visual autoregressive modeling},
  author={Qin, Ziran and Lv, Youru and Lin, Mingbao and Guo, Hang and Zhang, Zeren and Zou, Danping and Lin, Weiyao},
  booktitle={Proceedings of the AAAI Conference on Artificial Intelligence},
  volume={40},
  number={30},
  pages={24982--24990},
  year={2026}
}

@article{xiang2025make,
  title={Make it efficient: Dynamic sparse attention for autoregressive image generation},
  author={Xiang, Xunzhi and Fan, Qi},
  journal={arXiv preprint arXiv:2506.18226},
  year={2025}
}

%
%
%
%
%






\newpage 

\appendix
\section{Artifact Appendix}

\subsection{Abstract}
\noindent
Our artifact provides scripts to reproduce the image quality, latency, and throughput results of diagonal sparse attention and the baseline methods.
It contains the source code and a README that describes the exact commands to build the environment, run the experiments, and regenerate Figures 15–17 and Table 2 from the paper.

\subsection{Artifact check-list (meta-information)}

{\small
\begin{itemize}
  \item {\bf Algorithm: } Diagonal-aware sparse attention algorithm.
  \item {\bf Compilation: } CUDA toolkit, PyTorch, Triton
  \item {\bf Model: } Janus-Pro-1B, Janus-Pro-7B, Lumina-mGPT-7B-512
  \item {\bf Data set: } GenEval, DPG-Bench
  \item {\bf Run-time environment: } Docker with NVIDIA Container Toolkit; base image: nvcr.io/nvidia/pytorch:23.12-py3
  \item {\bf Hardware: } NVIDIA RTX A6000(48GB), Intel Xeon Gold. Evaluators need Ampere-or-newer GPU with at least 48GB memory. 
  \item {\bf Execution: } End-to-end bash scripts; environment initialization, experiments, figure generation.
  \item {\bf Metrics: } Image quality(GenEval score, DPG-bench score), Latency, Throughput(img/s)
  \item {\bf Output: } Generated images, figures and table.
  \item {\bf Experiments: } Figure 15, 16, 17 and Table 2 reported in the paper.
  \item {\bf How much disk space required (approximately)?: } 80GB free storage.
  \item {\bf How much time is needed to prepare workflow (approximately)?: } 1 hr
  \item {\bf How much time is needed to complete experiments (approximately)?: } 1 week for quality experiments(Table 2), 1.5 hr for speedup experiment(Fig 15-17).
  \item {\bf Publicly available?: } Yes
  \item {\bf Code licenses (if publicly available)?: } Creative Commons Attribution 4.0 International, MIT License
  \item {\bf Archived (provide DOI)?: } 
  
  https://doi.org/10.5281/zenodo.21712898
\end{itemize}
}

\subsection{Description}

\subsubsection{How to access}

Clone our public Github repository: 

\begin{verbatim}
$ git clone \
https://github.com/casys-kaist/DiagonalAttn
\end{verbatim}


\subsubsection{Software dependencies}

We distribute the artifact as a Docker image, so evaluators need Docker and the NVIDIA Container Toolkit.
We assume the host has CUDA 12.3, matching the version used by the provided base image.
Python dependencies are installed with \texttt{uv}, using a provided initialization script.

\subsubsection{Datasets}

The prompt datasets for GenEval and DPG-bench are already included in the repository.

\subsubsection{Models}

We provide a helper script that automatically downloads all required model weights.

\subsection{Installation}

\noindent \circleN{1}~\textbf{Set up the Docker Container.}
Before launching the Docker container, set the correct \texttt{HOST\_CACHE\_DIR} path in \texttt{launch.sh}.
\begin{verbatim}
$ cd DiagonalAttn/docker 
$ sh launch.sh
\end{verbatim}

\noindent \circleN{2}~\textbf{Initialize environment.}
After attaching to the running container, set the environment variables and install the required packages.
\begin{verbatim}
$ source /workspace/setup/env.sh
$ source /workspace/setup/uv_env.sh
\end{verbatim}

\noindent \circleN{3}~\textbf{Download models.}
Download the model weights from Hugging Face.
Evaluators should log in to Hugging Face before downloading.
\begin{verbatim}
$ hf auth login 
$ cd /workspace
$ bash setup/download_models.sh
\end{verbatim}

\subsection{Experiment workflow}

\niparagraph{Quality Experiment.}
This experiment reproduces Table 2, comparing image quality across the baseline methods and diagonal attention.
The scripts first generate images for the baseline methods and diagonal attention.
Then, it aggregates the score results written in \texttt{/workspace/results} and render them into table.  
Evaluating all prompts(\texttt{-{}-ratio 1.0}) for quality takes multiple GPU-weeks; it may take $\sim$40 days on 4 RTX A6000 GPUs.
We provide \texttt{-{}-ratio} to control the fraction of prompts evaluated.
The results reported in the paper used \texttt{-{}-ratio 1.0}.

%
\begin{verbatim}
$ bash evaluation/quality/run_table2.sh \
       --all-cells --gpus 0,1,2,3 --ratio 0.1
$ python evaluation/quality/render_table2.py
\end{verbatim}

\niparagraph{Speedup Experiment.}
Each script generates a PDF for its corresponding figure among Figures 15, 16, and 17.
For Figure 15, \texttt{fig15\_quality.sh} must be run first to produce the quality-axis results, which \texttt{fig15.sh} then joins with the latency-axis measurements.
The script then measures the speedup and automatically renders the figure as a PDF.
\begin{verbatim}
$ bash speedup/figure/fig15_quality.sh \
       --gpus 0,1,2,3 --ratio 0.1
$ LOCK_GPU_CLOCKS=1 bash speedup/figure/fig15.sh
$ LOCK_GPU_CLOCKS=1 bash speedup/figure/fig16.sh
$ LOCK_GPU_CLOCKS=1 bash speedup/figure/fig17.sh
\end{verbatim}

\subsection{Evaluation and expected results}
\noindent
After running the scripts, the artifact generates the table and figures under \texttt{./results}, matching the expected results below:
\begin{verbatim}
◦ Table 2
   - results/table2.md
◦ Figure 15
   - results/figures/fig15.pdf
◦ Figure 16
   - results/figures/fig16a.pdf
   - results/figures/fig16b.pdf
◦ Figure 17
   - results/figures/fig17.pdf
\end{verbatim}
The PDF figures and Markdown table correspond to the plots and table reported in the paper.
In addition to these final outputs, each experiment also stores raw logs under \texttt{results/logs} and \texttt{results/speedup}, and image outputs under \texttt{results/images}.

\subsection{Experiment customization}
\noindent
We provide several input parameters for customizing the experiments.

\noindent \textbf{\texttt{evaluation/quality/run\_table2.sh}}
\begin{itemize}
  \item {\texttt{-{}-gpus}: } CUDA GPU IDs to round-robin parallel execution of each cells.
  \item {\texttt{-{}-models}: } Restrict to specific model (\texttt{januspro1b}, \texttt{janus7b}, \texttt{lumina}).
  \item {\texttt{-{}-benchmarks}: } Restrict to specific benchmarks (\texttt{geneval}, \texttt{dpg}).

  \item {\texttt{-{}-methods}: } Restrict to specific attention methods (\texttt{dense}, \texttt{alisa}, \texttt{h2o}, \texttt{tova}, \texttt{window}, \texttt{streamingllm}, \texttt{diagonal}, \texttt{diagonal\_sink4}).
  \item {\texttt{-{}-sparsity}: } Restrict to specific sparsity levels (e.g.\ \texttt{0.950}).
  \item {\texttt{-{}-ratio}: } Override the benchmark prompt ratio for reduced runtime.
\end{itemize}

\noindent \textbf{\texttt{evaluation/quality/render\_table2.py}}
\begin{itemize}
  \item {\texttt{-{}-ratio}: } Select which recorded benchmark-ratio tier's rows to render.
  \item {\texttt{-{}-precision N}: } Number of decimal places shown in the rendered table.
  \item {\texttt{-{}-out-md, -{}-out-tex}: } Output table Markdown/LaTex table paths.
\end{itemize}

\noindent \textbf{\texttt{speedup/figure/fig15\_quality.sh}}
\begin{itemize}
  \item {\texttt{-{}-gpus}: } CUDA GPU IDs to parallelize the GenEval quality cells across.
  \item {\texttt{-{}-ratio}: } Subsample GenEval prompts to for reduced runtime.
\end{itemize}

\subsection{Notes}
\noindent
More information can be found in the \texttt{README} file of each directory.






\end{document}